\documentclass[12pt]{article}
\usepackage[margin=1in]{geometry}
\usepackage[T1]{fontenc}
\usepackage[scaled]{helvet}

\usepackage{amsmath,amssymb,amsthm}
\usepackage{bm}
\usepackage{graphicx}
\usepackage{booktabs}
\usepackage{tabularx}
\usepackage{array}
\usepackage{xcolor}
\usepackage[numbers,sort&compress,square]{natbib}
\usepackage[colorlinks=true,linkcolor=blue!50!black,citecolor=blue!50!black,urlcolor=blue!50!black]{hyperref}
\usepackage[capitalise]{cleveref}
\usepackage{caption}
\crefname{figure}{Fig.}{Figs.}
\Crefname{figure}{Fig.}{Figs.}
\crefname{section}{Section}{Sections}

\newtheorem{proposition}{Proposition}

\newcommand{\R}{\mathbb{R}}
\newcommand{\model}{NOMAD-EEG}

\begin{document}

\begin{titlepage}
\begin{center}
{\Large\bfseries NOMAD-EEG: Native Omni-Montage Adaptive Diffusion\par}
\vspace{1.5em}
{\large Abdulkader Helwan$^{1}$, Khadidja Henni$^{2}$ and Lina Abou-Abbas$^{1,*}$\par}
\vspace{1em}
{\small
$^{1}$Department of Electrical and Computer Engineering, Lebanese American University, Byblos, Lebanon\\[0.3em]
$^{2}$Institute of Applied Artificial Intelligence, T\'ELUQ University, Montreal, Canada\\[0.3em]
$^{*}$Author to whom any correspondence should be addressed: \href{mailto:lina.abouabbas@lau.edu.lb}{lina.abouabbas@lau.edu.lb}\par}
\vspace{1em}

\end{center}
\vspace{2em}

\end{titlepage}

\section*{Abstract}
\noindent\textbf{Purpose.} Electroencephalography (EEG) montages, channel counts and recording lengths
vary widely, but most EEG foundation models are bound to the channel set and input length they were
trained on. We aimed to build a generative EEG foundation model independent of the number, identity
and order of electrodes and of the input length.

\noindent\textbf{Methods.} \model{} is a 12.0\,M-parameter diffusion model that alternates
state-space layers along time, without patches, with attention across
electrodes identified only by their 3D scalp position. It was pre-trained on 147\,h of unlabelled EEG
from a single corpus and evaluated with frozen linear probes on ten datasets with 1 to 64 electrodes
against seven foundation models run under one protocol, and on two unseen populations.

\noindent\textbf{Results.} Despite its small size (12.0\,M parameters against 69\,M for REVE) and
single pre-training corpus, \model{} outperformed all seven foundation models on EEGMMIDB motor
imagery (balanced accuracy 0.442), Siena seizure detection (macro F1 0.622) and Bonn seizure
detection (AUROC 0.999), had the highest accuracy of the seven compared models on TUEP epilepsy
diagnosis (0.640) and, after fine-tuning, the highest $\kappa$ (0.635) and weighted F1 (0.814) on TUEV
event classification. It was the best of five models on P300 detection in autism with an unseen
eight-electrode montage (AUROC 0.718). With only 5 of 19 electrodes, its mean AUROC fell only from
0.942 to 0.891.

\noindent\textbf{Conclusion.} Identifying electrodes by position and modelling time without patches
let one small model, pre-trained on one corpus, outperform much larger EEG foundation models on
seizure detection, epileptiform event classification, epilepsy diagnosis and motor imagery.

\medskip
\noindent\textbf{Keywords:} EEG foundation model; Diffusion model; Montage-agnostic; State-space model;
Self-supervised learning; Seizure detection

\newpage

\section{Introduction}
\label{sec:intro}

Electroencephalography (EEG) is recorded with very different hardware and conventions.
A clinical epilepsy study may use the 19--21 electrodes of the international 10--20 system,
often re-referenced into a bipolar montage; a motor-imagery study may use a 64-electrode
10--10 cap; a sleep study may record only two bipolar derivations; and some archives hold a
single intracranial or scalp channel. Recording lengths vary just as widely, from 4\,s trials
in brain--computer interfaces to the 30\,s epochs of sleep scoring. At the same time, the
events clinicians care about are brief: a spike lasts 20--70\,ms and a sharp wave
70--200\,ms~\citep{kane2017glossary}. An EEG foundation model is only broadly useful if one
set of pretrained weights can read all of these inputs, at the temporal resolution the
events require.

Most EEG foundation models are, however, bound to the format they were trained on.
Some identify each channel by a learned embedding looked up in a fixed vocabulary of
electrode names or slots, so channels outside that vocabulary cannot be represented
(LaBraM~\citep{jiang2024labram}, EEGPT~\citep{wang2024eegpt}, BIOT~\citep{yang2023biot}).
Others mix channels with convolutions along the channel index, so the result depends on
the order in which channels are listed (CBraMod~\citep{wang2025cbramod},
EEGMamba~\citep{wang2025eegmamba}). Most tokenise time into non-overlapping patches of
about one second, and some cap the number of patches, which fixes a maximum input length.

Generative pre-training has recently emerged as a strong alternative to masked
reconstruction for EEG. Denoising diffusion~\citep{ho2020ddpm} was used by our earlier
DiffEEG model~\citep{diffeeg2026}; EEGDM~\citep{puah2025eegdm} combined diffusion with a
structured state-space backbone; and B[FM]$^2$~\citep{hwang2026bfm} showed that flow
matching~\citep{lipman2023flow} on the raw waveform reaches state-of-the-art accuracy on
seven of nine downstream tasks while pre-training on only about 307\,h of EEG. Yet each of
these generative models inherits a fixed input format. DiffEEG maps its input into a fixed
22-channel layout, EEGDM conditions a single-channel network on one of 22 bipolar channel
identities, and B[FM]$^2$ convolves along the electrode index of what it describes as a
small, fixed set of electrodes.

Montage-agnostic models do exist, but they take the opposite design choices.
REVE~\citep{ouahidi2025reve}, LUNA~\citep{doner2025luna} and
BrainOmni~\citep{xiao2025brainomni} place every channel by its 3D sensor position and so
accept arbitrary electrode sets. All three are trained with masked or tokenised
reconstruction objectives on transformers that operate on temporal patches or windows,
which trades temporal resolution for tractable attention. No existing model, to our
knowledge, combines a generative objective, position-based montage independence and
patch-free temporal modelling (\cref{tab:related}).

\begin{figure}[t]
  \centering
  \includegraphics[width=\textwidth]{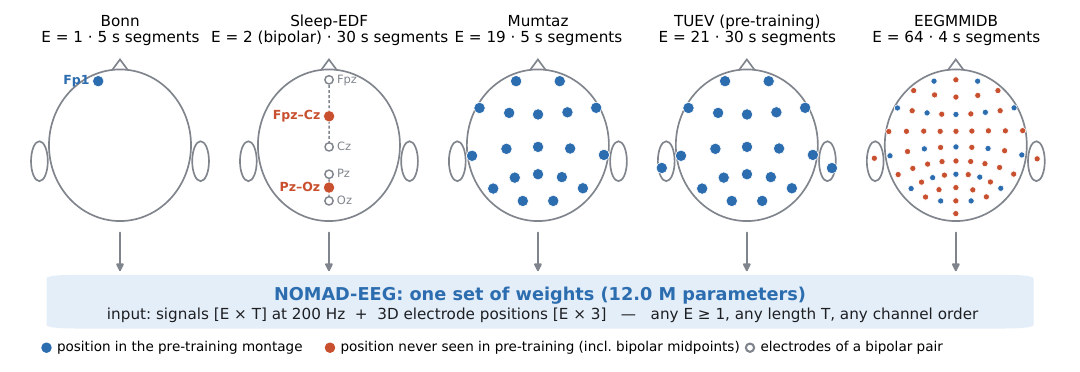}
  \caption{\textbf{One model, any montage.} The same pretrained \model{} weights read five
  electrode layouts used in this work, from a single electrode to a 64-electrode cap. Electrode
  positions are the 3D coordinates the network receives, projected to 2D (nose up). Blue:
  positions present in the pre-training montage. Red: positions never seen in pre-training,
  or the midpoint used for a bipolar derivation (open circles mark its two electrodes).}
  \label{fig:montages}
\end{figure}

We introduce \model{} (\emph{Native Omni-Montage Adaptive Diffusion}), a diffusion
foundation model whose architecture has no parameter tied to the number, identity or
order of electrodes, and none tied to the input length (\cref{fig:montages}). Every block
of the network factorises spatio-temporal mixing into two operations. A bidirectional
diagonal state-space layer (S4D)~\citep{gu2022s4d} mixes information along time within each
electrode, at a fixed 20\,ms resolution and with a convolution kernel generated for
whatever length the input has. Multi-head attention then mixes information across
electrodes within each time step, where each electrode is identified only by Fourier
features~\citep{tancik2020fourier} of its 3D scalp coordinate. The network is trained to
denoise EEG under a cosine noise schedule~\citep{nichol2021iddpm}, and its gated
intermediate activations are read out as representations.

Our contributions are:
\begin{itemize}
  \item \textbf{A montage- and length-agnostic generative backbone.} \model{} is, among the
  models we review, the only EEG foundation model that is simultaneously diffusion-pretrained,
  places electrodes by 3D position in every block, and models time without patches at 20\,ms
  resolution with no maximum length, pre-trained on 30\,s windows (1{,}500 steps per electrode)
  with a receptive field spanning the whole input. We show that its outputs are permutation-equivariant in
  the electrodes and that its parameter count is independent of the number of electrodes and
  samples (\cref{sec:properties}).
  \item \textbf{Task heads that keep this independence.} We define a frozen feature extractor
  (windowed mean and standard deviation of gate activations) and a token-based fine-tuning
  head whose output size does not depend on the montage or the recording length
  (\cref{sec:readouts}).
  \item \textbf{A controlled comparison of diffusion and flow matching.} Because the objective
  is a single switch in our implementation, we pre-train two models that differ only in the
  objective and compare them on every downstream task.
  \item \textbf{A single-protocol benchmark.} We re-run seven published EEG foundation models
  from their released weights on eight datasets under one evaluation protocol, rather than
  quoting results obtained under different splits, and add transfer to two populations unseen in
  pre-training (P300 detection in autism and Parkinson's disease diagnosis).
  \item \textbf{Ablations and a scaling study.} We measure what pre-training adds over random
  weights, test electrode-order invariance and missing electrodes, and train a 5.6-times larger
  model on 47 times more data to test whether scale helps.
\end{itemize}

Pre-trained on 147\,h of unlabelled TUEV recordings, \model{} reads datasets with one to 64
electrodes with a single set of weights. Across eight benchmarks it is never ranked below fourth
among eight foundation models and is first on Bonn and EEGMMIDB, and it is the best of five models on
P300 detection with an eight-electrode montage it never saw. A larger model trained on far more data
was not significantly better on any benchmark.

\subsection{Related work}
\label{sec:related}

\paragraph{EEG foundation models.}
Self-supervised pre-training on large unlabelled EEG corpora, most often the Temple
University Hospital (TUH) EEG corpus~\citep{obeid2016tuh}, has become the dominant route to
transferable EEG representations~\citep{kuruppu2025review,zhou2025survey}. BENDR~\citep{kostas2021bendr}
adapted contrastive speech pre-training to EEG. BIOT~\citep{yang2023biot} tokenises each
channel's short-time spectrum and adds a learned channel token, which lets one model read
datasets with different channel counts up to the size of its token table. LaBraM~\citep{jiang2024labram}
and EEGPT~\citep{wang2024eegpt} pre-train transformers on one-second patches (LaBraM) or
short windows (EEGPT) with masked objectives, identifying each electrode by a learned
embedding drawn from a fixed vocabulary of channel names. CBraMod~\citep{wang2025cbramod}
separates spatial and temporal attention in a criss-cross transformer and replaces channel
embeddings by a convolutional positional encoding computed over the channel\,$\times$\,patch
grid; EEGMamba~\citep{wang2025eegmamba} keeps the same input contract and replaces the
temporal transformer with Mamba~\citep{gu2023mamba}.

\paragraph{Montage-agnostic EEG models.}
Several recent models remove the fixed channel vocabulary. MMM~\citep{yi2023mmm} maps every
montage onto a unified topology of region tokens using electrode geometry.
REVE~\citep{ouahidi2025reve}, pre-trained on about 60{,}000\,h of EEG, adds a 4D Fourier
positional encoding of each electrode's 3D coordinate and patch index, and applies full
attention over all channel--patch tokens. LUNA~\citep{doner2025luna} encodes 3D coordinates
with a NeRF-style~\citep{mildenhall2020nerf} sinusoidal projection and compresses any number of
channels into a few learned queries by cross-attention before temporal attention, so its
cost grows linearly with channel count. BrainOmni~\citep{xiao2025brainomni} encodes sensor
position, orientation and type to unify EEG and MEG in a discrete tokeniser.
DIVER-0~\citep{han2025diver0} achieves channel-permutation equivariance without coordinates,
using attention biases to separate channels. Complementary work adapts fixed-montage models to
new channel sets after pre-training, for example by interpolation or learned projections
\citep{channeladapt2026}. Compatibility with a new montage does not guarantee robustness to it:
\citet{steglich2026montage} found that montage-agnostic models such as LUNA and REVE can still
lose accuracy when the electrode set changes. All of these models are trained with masked or
tokenised reconstruction objectives and operate on temporal patches or windows.

\paragraph{Generative pre-training for EEG.}
Diffusion models~\citep{ho2020ddpm,nichol2021iddpm} and flow
matching~\citep{lipman2023flow,liu2023rectified} learn to reverse a noising process and have
recently been used as EEG pre-training objectives. Our earlier DiffEEG~\citep{diffeeg2026}
pre-trained a U-Net to denoise 22-channel EEG and read features from its encoder.
EEGDM~\citep{puah2025eegdm} pre-trains a WaveNet-style~\citep{oord2016wavenet} network of
structured state-space layers as a diffusion model; it processes one channel at a time,
conditioned on one of 22 bipolar channel identities, and reads gated activations as features.
B[FM]$^2$~\citep{hwang2026bfm} pre-trains a SplitUNet velocity network with flow matching on
the raw waveform. Its ablation shows that electrodes should interact in every block and that
factorising each block into a temporal and an electrode operation outperforms joint
two-dimensional convolution. SplitUNet is electrode-count-invariant, but its electrode
convolution treats neighbouring channel indices as neighbouring electrodes, and it
down-samples time by a factor of eight. TGSD~\citep{tgsd2026} combines state-space models with
diffusion for a different goal, EEG spatial super-resolution, rather than representation
learning. \model{} adopts the factorised block of B[FM]$^2$ and the gate-activation features of EEGDM,
but replaces index-based electrode mixing with coordinate-conditioned attention and replaces
temporal down-sampling with a full-resolution state-space layer.

\paragraph{State-space sequence models.}
Structured state-space models (S4)~\citep{gu2022s4} model long sequences through a linear
recurrence that can be evaluated as a global convolution. The diagonal variant
S4D~\citep{gu2022s4d} makes the kernel a sum of complex exponentials that can be generated in
closed form for any sequence length. Selective state-space models such as
Mamba~\citep{gu2023mamba} make the recurrence input-dependent. We use S4D because its
kernel is defined for every length without a learned position table, which is what makes
the time axis of \model{} length-agnostic.

\begin{table}[!htbp]
\centering
\caption{How EEG foundation models identify channels and handle time. Entries for the eight
models we run were checked against their released code; the rest are from the papers.
``Any electrode set'' means the model accepts channels that were not in its pre-training
montage without retraining.}
\label{tab:related}
\footnotesize
\setlength{\tabcolsep}{3.5pt}
\begin{tabularx}{\textwidth}{@{}l l >{\raggedright\arraybackslash}X l l >{\raggedright\arraybackslash}X@{}}
\toprule
Model & Objective & Channel identity & Any set? & Order-free? & Time handling \\
\midrule
\textbf{\model{} (ours)} & Diffusion & 3D coordinates (Fourier), every block & Yes & Yes & 20\,ms steps, no patches, no length cap \\
REVE~\citep{ouahidi2025reve} & Masked AE & 3D coordinates + patch index & Yes & Yes & 1\,s patches, full attention \\
LUNA~\citep{doner2025luna} & Masked recon. & 3D coordinates (NeRF-style) & Yes & Yes & Patches; channels compressed to queries \\
BrainOmni~\citep{xiao2025brainomni} & Tokeniser + masked & Sensor position, orientation, type & Yes & Yes & Windows; channels compressed \\
DIVER-0~\citep{han2025diver0} & Masked recon. & Attention bias, no coordinates & Yes & Yes & 1\,s patches \\
LaBraM~\citep{jiang2024labram} & Masked codes & Learned embedding per name & Vocabulary & Yes & 1\,s patches, $\le$16 patches \\
EEGPT~\citep{wang2024eegpt} & Masked, aligned & Learned embedding per channel & Vocabulary & Yes & Short windows \\
BIOT~\citep{yang2023biot} & Superv./contrast. & Learned token per slot & $\le$18 slots & No & STFT tokens, any length \\
CBraMod~\citep{wang2025cbramod} & Masked recon. & Conv.\ encoding over channel axis & Any count & No & 1\,s patches \\
EEGMamba~\citep{wang2025eegmamba} & Masked recon. & As CBraMod & Any count & No & 1\,s patches, Mamba \\
EEGDM~\citep{puah2025eegdm} & Diffusion & One of 22 bipolar channel IDs & No & ---$^{a}$ & 5\,s chunks \\
B[FM]$^2$~\citep{hwang2026bfm} & Flow matching & Conv.\ along electrode index & Any count & No & Time down-sampled $8\times$ \\
DiffEEG~\citep{diffeeg2026} & Diffusion & Fixed input slot & No & No & Global pooling over time \\
\bottomrule
\end{tabularx}
\\[2pt]
\raggedright\footnotesize $^{a}$ EEGDM processes channels independently, so order does not
matter, but channels do not interact in the backbone.
\end{table}

\section{Materials and Methods}
\label{sec:methods}

\subsection{Problem setting and input representation}
\label{sec:input}

An EEG recording is a matrix $\mathbf{x}\in\R^{E\times T}$ of $E$ channels and $T$ samples,
together with a position $\mathbf{p}_e\in\R^{3}$ for every channel $e$. Neither $E$ nor $T$
is fixed, and channels may be listed in any order. The model sees the pair
$(\mathbf{x},\mathbf{P})$, with $\mathbf{P}=[\mathbf{p}_1,\dots,\mathbf{p}_E]^\top$, and
never a channel index.

\paragraph{Electrode positions.}
Positions are looked up by channel name in the standard 10--20 and 10--05 montages of
MNE-Python~\citep{gramfort2013mne}; where the two systems overlap, the 10--20 position is
used. A bipolar derivation ``A--B'' is placed at the midpoint of A and B. All positions are
divided by one fixed constant, the largest norm, in MNE's head coordinate frame, among the 21
electrodes of the pre-training montage. Because this constant does not depend on the
montage being read, an electrode receives the same coordinate whether it appears in a
19-electrode or a 64-electrode recording. Channels without a standard position (for example
ECG, EMG, and the ROC channel of TUH recordings) are dropped rather than assigned an
invented position.

\paragraph{Signal preprocessing.}
All signals are resampled to 200\,Hz. Each channel is normalised with statistics computed
per corpus on training data only: the median is subtracted, the result is divided by the
inter-quartile range, and values are clipped to $\pm 20$. This robust scaling absorbs
differences in units and amplitude conventions between corpora while limiting the influence
of large artefacts. Each corpus also receives an integer identifier $c$ that conditions the
network (\cref{sec:block}).

\subsection{Architecture}
\label{sec:arch}

\begin{figure}[!tp]
  \centering
  \includegraphics[width=\textwidth]{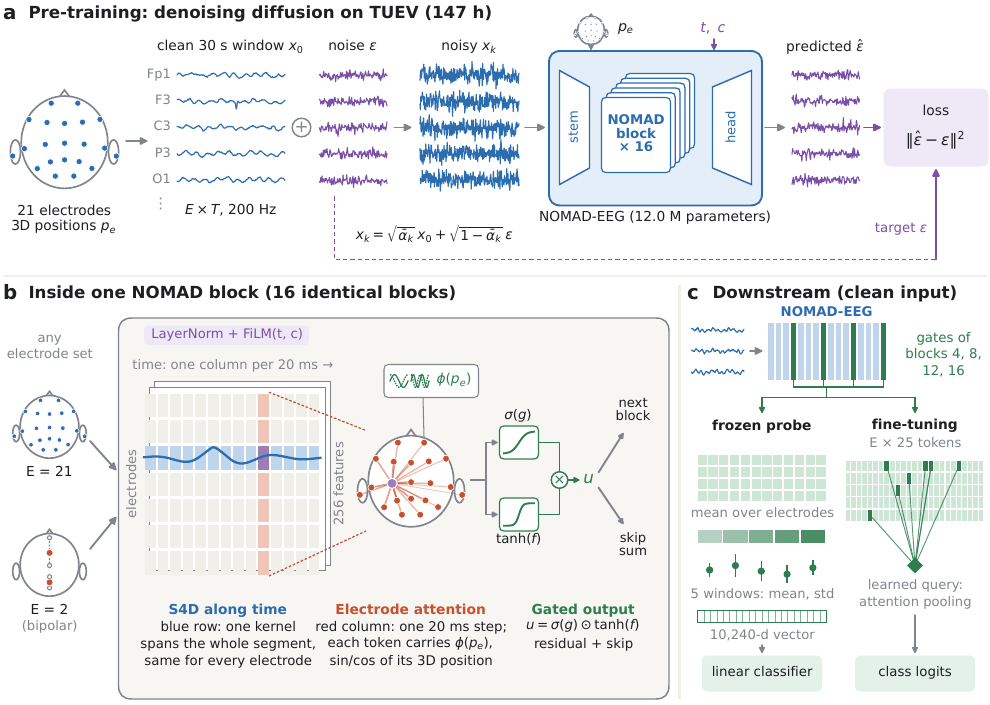}
  \caption{\textbf{Overview of \model{}.} (a)~Pre-training. A clean 30\,s TUEV window
  $\mathbf{x}_0$ (21 electrodes, 200\,Hz) is corrupted with Gaussian noise $\bm{\epsilon}$
  following the cosine schedule, and the network, given the electrode positions
  $\mathbf{p}_e$, the noise level $t$ and the corpus $c$, predicts the noise. It consists of a
  convolutional stem shared by all electrodes (one 256-dimensional feature every 20\,ms), 16
  identical blocks, and an output head that maps the summed skip outputs back to the sample
  rate. The flow-matching variant differs only in the objective. (b)~One block. The signal is
  a grid of tokens with one row per electrode and one column per 20\,ms step. After layer
  normalisation and FiLM conditioning on $(t,c)$, a bidirectional S4D layer filters each row
  along time with a kernel that spans the whole segment and is shared by all electrodes
  (blue); attention then mixes each column across electrodes, every token carrying Fourier
  features $\bm{\phi}(\mathbf{p}_e)$ of its electrode's 3D position (red). Because electrodes
  are identified only by position, the same block reads any electrode set, for example the
  21-electrode pre-training montage or two bipolar channels placed at the midpoints of their
  electrode pairs. A gated activation gives the output $\mathbf{u}$, which feeds the next block
  and the skip sum. (c)~Downstream, the input is clean and the gates of blocks 4, 8, 12 and 16
  are read. The frozen probe averages them over electrodes and keeps the mean and standard
  deviation in five time windows (10{,}240 features) for a linear classifier; the fine-tuning
  head pools $E\times 25$ tokens by attention from a learned query (\cref{fig:readouts}).
  Waveforms, attention weights and highlighted tokens are illustrative; electrode positions are
  the coordinates the network receives.}
  \label{fig:arch}
\end{figure}

\model{} (\cref{fig:arch}) consists of a temporal stem, a stack of $N=16$ identical blocks
with residual and skip connections, and an output head used only for pre-training. The
feature width is $d=256$ throughout, and the model has 11{,}998{,}977 parameters.

\subsubsection{Stem}
Each channel is processed independently by a single one-dimensional convolution with one
input channel, $d$ output channels, kernel size 9, stride 4 and padding 4, whose weights are
shared by all electrodes:
\begin{equation}
  \mathbf{h}^{(0)}_{e} = \mathrm{Conv1d}(\mathbf{x}_e) \in \R^{T'\times d},
  \qquad T'=\lceil T/4\rceil .
\end{equation}
At 200\,Hz one step of $\mathbf{h}^{(0)}$ covers 20\,ms. The stem is the only temporal
down-sampling in the network, so every block operates at this resolution.
The result is a tensor $\mathbf{H}\in\R^{E\times T'\times d}$: one row per electrode and one
column per 20\,ms step (\cref{fig:factorised}).

\subsubsection{Conditioning}
The noise level $t\in[0,1]$ is embedded with sinusoidal features of $1000\,t$ and a
two-layer MLP with SiLU activation; a learned embedding of the corpus identifier $c$ is added:
\begin{equation}
  \mathbf{k} = \mathrm{MLP}\big(\gamma(1000\,t)\big) + \mathbf{w}_{c} \in \R^{d},
\end{equation}
where $\gamma$ is the standard sinusoidal embedding. Downstream, inputs are clean and $t$ is
set to the clean end of the noise schedule (\cref{sec:readouts}).

\subsubsection{The NOMAD block}
\label{sec:block}

\begin{figure}[t]
  \centering
  \includegraphics[width=\textwidth]{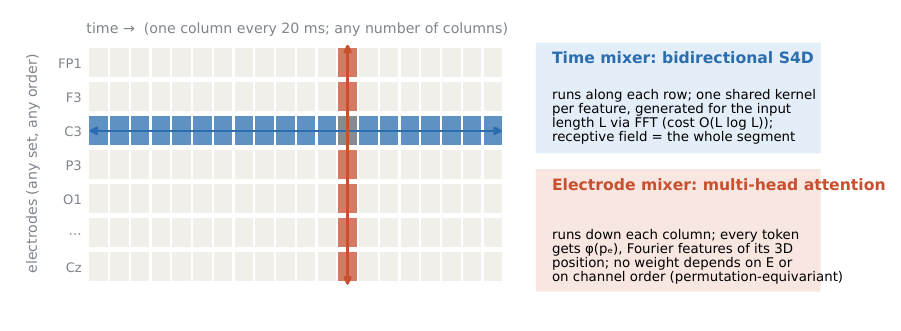}
  \caption{\textbf{Factorised mixing.} Inside the network an EEG segment is a grid with one
  row per electrode and one column per 20\,ms step. Each block mixes information along every
  row with a bidirectional S4D layer and then down every column with attention across
  electrodes. Neither operation has a parameter whose size depends on the number of rows or
  columns, so the grid can have any shape.}
  \label{fig:factorised}
\end{figure}

Following the finding of \citet{hwang2026bfm} that electrodes should interact in every block
and that factorised temporal and spatial mixing outperforms joint two-dimensional mixing,
each block applies one temporal and one spatial operation (\cref{fig:factorised}). Given the
input $\mathbf{H}$ of block $n$, the block computes the following.

\emph{FiLM conditioning.} After layer normalisation, features are scaled and shifted by
functions of the conditioning vector~\citep{perez2018film}:
\begin{equation}
  \mathbf{Z} = \mathrm{LN}(\mathbf{H})\odot\big(1+\bm{\alpha}(\mathbf{k})\big)+\bm{\beta}(\mathbf{k}),
\end{equation}
where $[\bm{\alpha};\bm{\beta}]$ is a linear map of $\mathbf{k}$ initialised to zero, so each
block starts as an unconditioned residual block.

\emph{Temporal mixing with S4D.} Each electrode's sequence $\mathbf{Z}_{e}\in\R^{T'\times d}$
is filtered, feature by feature, by a bidirectional diagonal state-space layer
\citep{gu2022s4,gu2022s4d}. For feature channel $j$, the forward kernel of length $L=T'$ is
\begin{equation}
  K^{\rightarrow}_{j}[\ell] \;=\; 2\,\mathrm{Re}\sum_{n=1}^{N_s}
     C_{jn}\,\frac{e^{\Delta_j A_{jn}}-1}{A_{jn}}\,e^{\ell\,\Delta_j A_{jn}},
  \qquad \ell=0,\dots,L-1,
\end{equation}
the zero-order-hold discretisation of a diagonal continuous-time state-space model with
$N_s=32$ complex modes $A_{jn}$, output weights $C_{jn}$ and step size $\Delta_j$. A second,
independently parameterised kernel $K^{\leftarrow}_j$ is applied to the time-reversed
sequence, and the two are combined into one non-causal convolution computed with the FFT:
\begin{equation}
  \mathbf{y}_{e,:,j} = \big(K^{\rightarrow}_j * \mathbf{z}_{e,:,j}\big)
     + \mathrm{flip}\big(K^{\leftarrow}_j * \mathrm{flip}(\mathbf{z}_{e,:,j})\big)
     + D_j\,\mathbf{z}_{e,:,j}.
\end{equation}
We use the S4D-Lin initialisation $A_{jn}=-\tfrac12+i\pi n$ and draw $\log\Delta_j$ uniformly
between $\log 10^{-3}$ and $\log 10^{-1}$. Because the kernel is a closed-form function of
$\ell$, it is generated for the length $L$ of each input; there is no learned position table
and no maximum length. The layer's receptive field is the whole segment from the first block,
and its cost is $O(E\,d\,L\log L)$.

\emph{Spatial mixing with coordinate-conditioned attention.} At each time step $\tau$ the $E$
electrode features $\{\mathbf{y}_{e,\tau}\}_{e=1}^{E}$ form a set of tokens. Each token is
augmented with Fourier features~\citep{tancik2020fourier} of its electrode's position,
\begin{equation}
  \bm{\phi}(\mathbf{p}) = \big[\sin(2^{k}\pi p_{a}),\,\cos(2^{k}\pi p_{a})\big]_{k=0,\dots,7;\;a\in\{x,y,z\}}
  \in\R^{48},
\end{equation}
projected linearly to $d$ dimensions, and the tokens attend to each other with four-head
self-attention~\citep{vaswani2017attention}:
\begin{equation}
  \mathbf{V}_{:,\tau} = \mathbf{Y}_{:,\tau} +
  \mathrm{MHA}\big(\mathbf{Y}_{:,\tau}+\mathbf{W}_{p}\bm{\phi}(\mathbf{P})\big).
\end{equation}
The positional features enter the attention input only; the residual stream carries the
signal features. Electrode identity is therefore defined by position alone, as in
REVE~\citep{ouahidi2025reve}, but here the positional information is re-injected in every
block rather than once at the input. Attention is computed independently at each of the
$T'$ steps, at a cost of $O(T'\,E^{2}d)$ per block.

\emph{Gated output.} A linear map produces two halves $[\mathbf{g};\mathbf{f}]$ from
$\mathbf{V}$, which are combined with the gated activation of
WaveNet~\citep{oord2016wavenet}, also used by EEGDM~\citep{puah2025eegdm}:
\begin{equation}
  \mathbf{U}=\sigma(\mathbf{g})\odot\tanh(\mathbf{f}),\qquad
  \mathbf{H}\leftarrow\mathbf{H}+\mathbf{W}_{r}\mathbf{U},\qquad
  \mathbf{S}\leftarrow\mathbf{S}+\mathbf{W}_{s}\mathbf{U}.
\end{equation}
The gate $\mathbf{U}^{(n)}\in\R^{E\times T'\times d}$ is bounded, which stabilises training
at depth, and it is the tensor from which downstream features are read.

\subsubsection{Output head}
The sum of skip outputs $\mathbf{S}$ over all blocks passes through layer normalisation,
SiLU and a linear layer, and a transposed convolution (kernel 8, stride 4) initialised to zero
maps each electrode back to the sample rate; a final linear interpolation matches the input
length $T$ exactly. The head is used only for the pre-training loss.

\subsection{Pre-training objective}
\label{sec:objective}

\model{} is pre-trained as a denoising diffusion model~\citep{ho2020ddpm} with the cosine
noise schedule of \citet{nichol2021iddpm} ($s=0.008$, 1{,}000 steps). For a clean window
$\mathbf{x}_0$, a step $k$ is drawn uniformly, the window is corrupted,
\begin{equation}
  \mathbf{x}_k=\sqrt{\bar\alpha_k}\,\mathbf{x}_0+\sqrt{1-\bar\alpha_k}\,\bm{\epsilon},
  \qquad \bm{\epsilon}\sim\mathcal{N}(\mathbf{0},\mathbf{I}),
\end{equation}
and the network, conditioned on $t=k/1000$, predicts the noise:
\begin{equation}
  \mathcal{L}_{\mathrm{diff}}=\mathbb{E}_{\mathbf{x}_0,k,\bm{\epsilon}}
  \big\|f_\theta(\mathbf{x}_k,t,\mathbf{P},c)-\bm{\epsilon}\big\|_2^2 .
\end{equation}
To predict the noise at every level, the network must model what EEG looks like: its
rhythms, transients and artefacts, and how signals co-vary across the scalp.

\paragraph{Flow-matching variant.}
To compare objectives under otherwise identical conditions, we also train the same network
with rectified flow~\citep{liu2023rectified,lipman2023flow}: $t\sim\mathcal{U}(0,1)$,
$\mathbf{x}_t=(1-t)\,\bm{\epsilon}+t\,\mathbf{x}_0$ and target $\mathbf{x}_0-\bm{\epsilon}$.
Architecture, data, optimiser, number of training epochs and downstream task heads are unchanged. We refer to the two
models as \model{} (diffusion, the default) and \model{}-F (flow).

\subsection{Montage and length independence}
\label{sec:properties}

The design gives \model{} two properties that do not depend on training.

\begin{proposition}[Electrode permutation equivariance]
Let $\Pi$ be any permutation of the $E$ electrodes. Every intermediate tensor, including the
gate activations $\mathbf{U}^{(n)}$ and the output, satisfies
$F(\Pi\mathbf{x},\Pi\mathbf{P})=\Pi\,F(\mathbf{x},\mathbf{P})$, and any output that pools
over electrodes is invariant to $\Pi$.
\end{proposition}
\emph{Reason.} The stem, layer normalisation, FiLM, S4D, gate and output projections act on
each electrode separately with shared weights and therefore commute with $\Pi$. Self-attention
without positional indices is permutation-equivariant, and the only electrode-specific input,
$\bm{\phi}(\mathbf{p}_e)$, moves with its electrode.

\begin{proposition}[Size independence]
No parameter of \model{} has a shape that depends on $E$ or $T$. The same weights therefore
apply to any number of electrodes $E\ge 1$, to electrodes absent from the pre-training
montage, and to any input length.
\end{proposition}
\emph{Reason.} The stem and the S4D parameters are per feature, the S4D kernel is generated
for the input length, and the attention and positional projection are per token.
Per block the cost is $O(E\,d\,T'\log T' + T'E^{2}d)$: roughly linear in time and quadratic in
the number of electrodes, which is small for scalp EEG.

\paragraph{Long context.}
Size independence also removes the usual ceiling on context length. \model{} is pre-trained on
30\,s windows, which are 1{,}500 steps per electrode at 20\,ms resolution, and every step can draw
on the whole window from the first block onwards, because each S4D kernel spans the entire
input. Longer recordings need no architectural change: the kernel is simply generated for the
longer length, and the cost grows as $O(L\log L)$ in the S4D layers and linearly in the
attention layers, which act on one time step at a time. By contrast, LaBraM's temporal embedding
covers at most 16 one-second patches~\citep{jiang2024labram}, and EEGDM folds inputs longer than
5\,s into independent 5\,s chunks~\citep{puah2025eegdm}, so neither relates events more than 16\,s
or 5\,s apart within one pass.
These properties guarantee that any montage and any length can be \emph{processed}, not that
every montage is represented equally well. Electrodes far from the pre-training positions rely
on the smoothness of $\bm{\phi}$, and robustness to montage changes has to be measured
separately~\citep{steglich2026montage}. The model still assumes 200\,Hz input and a known
position for every channel, and a bipolar derivation placed at its midpoint approximates, but
does not equal, a sensor at that point.

\subsection{Downstream task heads}
\label{sec:readouts}

\begin{figure}[t]
  \centering
  \includegraphics[width=\textwidth]{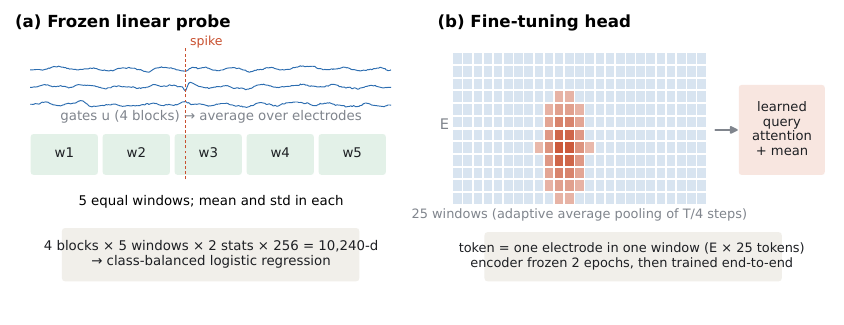}
  \caption{\textbf{Downstream task heads.} (a) The frozen feature extractor averages the gate activations
  over electrodes, splits time into five equal windows and keeps the mean and standard
  deviation in each, giving 10{,}240 features whatever the montage or length. (b) The
  fine-tuning head keeps electrodes separate: each electrode in each of 25 windows is a
  token, and a learned query pools the tokens by attention. Traces and attention shading are
  illustrative.}
  \label{fig:readouts}
\end{figure}

Downstream, the network receives clean EEG, and $t$ is set to the clean end of the
trajectory ($t=0$ for \model{} $t=1$ for \model{}-F); no noise is added. We read the gates
$\mathbf{U}^{(n)}$ of blocks $n\in\{4,8,12,16\}$. The generative output head of pre-training is not used.

\paragraph{Frozen feature extraction.}
For linear probing (\cref{fig:readouts}a) each gate is averaged over electrodes, which removes
the dependence on $E$. The $T'$ steps are divided into $W=5$ equal windows of
$\lfloor T'/5\rfloor$ steps (any remainder is dropped), and the mean and standard deviation of
each feature are taken within each window. Concatenating four blocks, five windows, two
statistics and 256 features gives a vector of 10{,}240 dimensions for every input. Windowed
statistics keep coarse temporal order and let brief events show up through the standard
deviation, which a single global average would dilute. Because the number of windows, not their
duration, is fixed, a 5\,s input gives 1\,s windows and a 30\,s input gives 6\,s windows. A
class-balanced linear classifier is trained on these vectors with the encoder frozen.

\paragraph{Fine-tuning head.}
For fine-tuning (\cref{fig:readouts}b) the four gates are concatenated to
$\R^{E\times T'\times 1024}$, and adaptive average pooling reduces time to $W=25$ windows per
electrode. Each window is projected by a linear layer to 256 dimensions, followed by GELU and
layer normalisation, and a learned embedding of the window index is added. The resulting
$E\times 25$ tokens, one per electrode per window, are pooled by four-head attention from a
single learned query; the pooled vector is concatenated with the mean of all tokens and mapped
to class logits by a linear layer after dropout (0.1). The head's parameters do not depend on
$E$ or $T$, so the same head applies to every dataset: 21 electrodes give 525 tokens and two
bipolar channels give 50. Training has two phases. For the first two epochs only the head is
trained, which prevents a randomly initialised head from disturbing the pre-trained encoder;
afterwards the whole network is trained, with learning rate $5\times10^{-4}$ for the head and
$5\times10^{-5}$ for the encoder, AdamW~\citep{loshchilov2019adamw} (weight decay 0.05),
5\% linear warm-up followed by cosine decay, and class-weighted cross-entropy. The checkpoint
with the best balanced accuracy on a subject-disjoint validation set is evaluated once on the
test set.

\subsection{Pre-training data and implementation}
\label{sec:impl}

\paragraph{Data.}
We pre-train on the TUH EEG Events corpus (TUEV)~\citep{harati2015tuev,obeid2016tuh},
using the recordings of both its training and evaluation partitions and no labels. Recordings
are read from the raw EDF files, reduced to the 21 channels of the 10--20 system that have
a standard position (ROC is dropped: it has no standard coordinate and is absent from 40\% of
TUEV samples), resampled to 200\,Hz with a polyphase filter, and cut into non-overlapping
30\,s windows that never cross a recording boundary. This yields 17{,}678 windows (147\,h),
of which 2\% are held out for validation. Thirty-second windows give the state-space layers
enough context to see rhythms evolve, and because the architecture is length-agnostic,
downstream inputs may be shorter or longer. Pre-training on TUEV means that TUEV downstream
results measure in-domain transfer; all other downstream datasets are external to
pre-training.

\paragraph{Optimisation.}
Both models are trained for about 74 epochs over the 17{,}325 training windows (40{,}000 weight
updates with an effective batch of 32 windows: 8 windows with 4-step gradient accumulation), with
AdamW (learning rate $3\times10^{-4}$, $\beta=(0.9,0.95)$, weight decay 0.01), a linear warm-up over
the first two epochs followed by cosine decay to zero, gradient
clipping at norm 1.0, and an exponential moving average of the weights with decay 0.999. The
moving-average weights are used for all downstream evaluations. Training uses bfloat16
autocasting, with the S4D FFT kept in float32 because the kernel tail underflows in bfloat16,
and gradient checkpointing~\citep{chen2016checkpoint} of each block. An update with a non-finite
loss or gradient is skipped. Each model trains on a single NVIDIA A100 GPU.

\subsection{Downstream datasets}
\label{sec:setup}

We evaluate on nine downstream datasets that differ in task, montage and segment length, compared
with all baselines, and on a tenth, the SEED-V emotion corpus, for \model{} alone
(\cref{tab:datasets}). TUAB, TUSZ, TUEV and TUEP come from the TUH EEG corpus~\citep{obeid2016tuh}; TUEV
was also part of \model{}'s unlabelled pre-training data, so its results measure in-domain
transfer. Siena, Bonn, Mumtaz, EEGMMIDB and Sleep-EDF are external to \model{}'s pre-training and
cover new hardware, montages and tasks; some of them were part of other models' pre-training data
(\cref{sec:baselines}). SEED-V is also external to \model{}'s pre-training. 

TUEP, the TUH Epilepsy Corpus, poses a diagnostic question: whether a patient has epilepsy, judged
from routine EEG of 100 patients with and 100 without epilepsy. Each patient contributes 60 random
5\,s windows. Because the two classes are equal in size, accuracy equals balanced accuracy and
weighted F1 equals macro F1 on this task. Under the SEED-V protocol of CBraMod the test set consists
of unseen film clips of the same subjects, and the validation clips are distinct from the training
clips.

\begin{table}[!htbp]
\centering
\caption{Downstream datasets. Train/val/test counts are segments; all splits are subject-disjoint
except TUAB, TUSZ, Siena and Bonn, which reuse the fixed splits of our earlier benchmark, and SEED-V,
which follows the CBraMod protocol (film clips 1--5, 6--10 and 11--15 of every session; all 16
subjects in every split). TUEP is evaluated by 5-fold cross-validation over its 200 patients.}
\label{tab:datasets}
\footnotesize
\setlength{\tabcolsep}{4pt}
\begin{tabular}{@{}l l c c c r@{}}
\toprule
Dataset & Task & Classes & Channels & Segment & Train / val / test \\
\midrule
TUAB & abnormal EEG & 2 & 21 & 5\,s & 100{,}000 / 20{,}000 / 50{,}000 \\
TUSZ & seizure & 2 & 21 & 5\,s & 100{,}000 / 20{,}000 / 50{,}000 \\
TUEV & EEG events & 6 & 21 & 5\,s & 83{,}932 (5-fold CV) / 29{,}421 \\
TUEP & epilepsy diagnosis & 2 & 19 & 5\,s & 12{,}000 from 200 patients (5-fold CV) \\
Siena & seizure & 2 & 21 & 5\,s & 74{,}512 / 8{,}279 / 18{,}729 \\
Bonn & seizure & 2 & 1 & 5\,s & 360 / 40 / 100 \\
Mumtaz & depression & 2 & 19 & 5\,s & 4{,}818 / 1{,}086 / 1{,}179 \\
EEGMMIDB & motor imagery & 4 & 64 & 4\,s & 6{,}300 / 1{,}734 / 1{,}803 \\
Sleep-EDF-20 & sleep stages & 5 & 2 (bipolar) & 30\,s & 29{,}996 / 3{,}175 / 9{,}137 \\
SEED-V & emotion & 5 & 59 & 1\,s & 34{,}432 / 42{,}960 / 40{,}352 \\
\bottomrule
\end{tabular}
\end{table}

\subsection{Baselines and overlap with their pre-training data}
\label{sec:baselines}

We compare \model{} with seven EEG foundation models, all run by us from their released weights:
EEGDM~\citep{puah2025eegdm}, REVE~\citep{ouahidi2025reve} (base), CBraMod~\citep{wang2025cbramod},
LaBraM~\citep{jiang2024labram} (base), BIOT~\citep{yang2023biot}, EEGMamba~\citep{wang2025eegmamba}
and our earlier DiffEEG~\citep{diffeeg2026}. Each model receives the same segments in its own native
input format (montage, sampling rate and amplitude scaling as in its authors' code) and keeps its
own feature pooling. Where a model's fixed channel vocabulary cannot represent a channel, the
channel is mapped to the nearest vocabulary entry by 3D position. B[FM]$^2$~\citep{hwang2026bfm}
could not be included because its code and weights have not been released.

Several evaluation datasets, or the corpus they were drawn from, appear in the pre-training data of
some models, according to each model's paper, README or released pre-training
code~\citep{puah2025eegdm,ouahidi2025reve,wang2025cbramod,jiang2024labram,diffeeg2026}. \model{} and
EEGDM were pre-trained on TUEV, BIOT's checkpoint on the TUAB and TUEV training splits, LaBraM on TUSZ,
TUEP, Siena and EEGMMIDB, and DiffEEG on the TUSZ training split. REVE, CBraMod and EEGMamba were
pre-trained on the TUH EEG corpus, from which TUAB, TUSZ, TUEV and TUEP are drawn, and REVE and EEGMamba
also on Siena; REVE removed the recordings of TUAB, TUEV, EEGMMIDB and Mumtaz from its pre-training
data. None of the models saw the labels of our test sets, but results on an overlapping dataset
measure in-domain rather than external transfer. Bonn, Mumtaz, Sleep-EDF and the BCIAUT-P300 data do
not appear in any model's documented pre-training data. LaBraM is not compared on TUEP because TUEP
was part of its pre-training data.

\subsection{Evaluation protocol}
\label{sec:protocol}

The main comparison is a \emph{frozen linear probe}. Encoders are frozen; a class-balanced,
L2-regularised logistic regression is trained on the training features, its regularisation strength
$C\in\{10^{-6},\dots,10^{3}\}$ is chosen on the validation split (by AUROC for binary and balanced
accuracy for multi-class tasks), and the test set is scored once. For TUEV we use \model{}'s own
probe (a softmax layer whose L2 strength and number of epochs are chosen by 5-fold subject-grouped
cross-validation) for every model. On TUEP the probe is evaluated by 5-fold cross-validation over
patients with identical folds for every model; its $C$ is chosen on 20\% of the training patients,
and a patient's score is the mean probability of their windows. We report a fixed set of metrics for
every dataset: AUROC, PR-AUC, balanced accuracy and macro F1 for binary tasks, and balanced accuracy,
Cohen's $\kappa$ and weighted F1 for multi-class tasks.

For TUEV, TUAB, TUSZ, Mumtaz, EEGMMIDB, Sleep-EDF and SEED-V we additionally fine-tune \model{}
end to end (\cref{sec:readouts}): two epochs with the encoder frozen, then 20 epochs end to end (10
for TUEV), the epoch chosen by validation balanced accuracy. The baselines were evaluated as frozen probes only, so a fine-tuned \model{} row is compared with
frozen baselines.\label{sec:finetune} For TUAB and TUSZ
the benchmark's validation windows are drawn from the training patients (the stored windows carry no
patient identifier), so fine-tuning there selects its epoch on patients it is trained on.

\subsection{Statistical analysis}
\label{sec:stats}

Every reported metric carries a 95\% confidence interval from 1{,}000 bootstrap resamples: of test
windows for the fixed-split benchmarks, and of subjects or patients for the cross-validated ones
(TUEP, BCIAUT-P300, Parkinson's disease). Differences between two models evaluated on the same test
samples (diffusion against flow matching, \model{} against a baseline on TUEP and Parkinson's disease)
are tested with a paired bootstrap on the same resamples (2{,}000 resamples for the objective
comparison), and a difference is called significant only
when its 95\% interval excludes zero. Where only unpaired intervals are available, we state whether
they overlap and do not call non-overlapping intervals a paired test.

\subsection{Transfer to unseen populations}
\label{sec:zeroshot}

To test whether \model{} transfers to a population, task and montage that none of the models
saw in pre-training, we evaluate frozen encoders on BCIAUT-P300~\citep{simoes2020bciaut} and on a
Parkinson's disease corpus. No encoder is trained or fine-tuned on any data from these corpora: the
pre-trained weights are used as they are, and only a linear classifier is fitted on top (a linear
probe), always on subjects other than the ones it is tested on.

\paragraph{BCIAUT-P300 data.}
BCIAUT-P300 contains EEG from 15 young people with autism spectrum disorder (ASD), each recorded
in 7 sessions of a P300-based brain--computer-interface training in which eight objects are
flashed and the participant attends to one of them~\citep{simoes2020bciaut}. EEG was recorded at
250\,Hz from eight electrodes (C3, Cz, C4, CPz, P3, Pz, P4, POz) and epoched from $-200$ to
1{,}200\,ms around each flash. Because every participant has ASD, the task is \emph{not} autism
diagnosis but P300 detection: deciding whether a flashed object was the attended target. We use
the labelled calibration data of every session: 20 blocks of 10 runs of 8 flashes, 1{,}600 epochs
per session and 168{,}000 epochs in total, of which 21{,}000 (12.5\%) are targets. Epochs are
resampled to 200\,Hz (280 samples, 1.4\,s) and normalised per channel with a median and
inter-quartile range pooled over all epochs without using labels, then clipped at $\pm20$.
The Kaggle release also contains labels for the online (test) sessions, which were withheld in
the original challenge; we do not use them.

\paragraph{Models and inputs.}
We compare \model{}-D with DiffEEG~\citep{diffeeg2026}, EEGDM~\citep{puah2025eegdm},
REVE~\citep{ouahidi2025reve} and BIOT~\citep{yang2023biot}, each fed in its native input format
(\cref{sec:related}). This montage is new to all five models, and they can read different
amounts of it. \model{} and REVE place all eight electrodes by their 3D positions; CPz and POz
are not in \model{}'s pre-training montage. DiffEEG's fixed 22-channel layout has slots for six
of the eight (C3, Cz, C4, P3, Pz, P4). EEGDM's vocabulary of bipolar channels allows four
derivations (C3--Cz, C4--Cz, C3--P3, C4--P4), and BIOT's allows two (C3--P3, C4--P4). Two models
also need an adaptation to the 1.4\,s epoch. EEGDM pools its gate activations over one window
spanning the whole epoch instead of 1\,s windows. REVE keeps its own 1\,s patches, so it reads a
single patch covering $-200$ to 800\,ms, which contains the P300 (typically 250--500\,ms after the
flash).

\paragraph{BCIAUT-P300 protocol.}
Features are extracted once with each frozen encoder (for \model{}, the frozen feature extractor of
\cref{sec:readouts}). The 15 subjects are split into five folds of three (subject-grouped
cross-validation). In each fold, a class-balanced, L2-regularised logistic regression is trained
on the other twelve subjects; its strength $C\in\{10^{-4},\dots,10\}$ is chosen by AUROC on three
of those twelve held out as validation, and the classifier is then refitted on all twelve and
applied to the three test subjects. Every epoch is therefore predicted exactly once, by a
classifier that never saw its subject. We report, on the pooled out-of-fold predictions,
AUROC, balanced accuracy and PR-AUC (chance level equals the target rate, 0.125), with 95\%
confidence intervals from 1{,}000 bootstrap resamples of subjects. We also report the decision a
P300 speller actually makes: in each block, the object whose flashes receive the highest mean
target probability over the first $k$ runs is selected, and we score how often it is the
attended object (chance 12.5\%) for $k=1$, 3 and 10.

\paragraph{Parkinson's disease.}
We apply the same unseen-participant protocol to a diagnostic task: Parkinson's disease (PD) against
healthy controls in OpenNeuro ds004584, 149 participants (100 PD, 49 controls) recorded at rest
with eyes open for about two minutes with a 64-channel cap at 500\,Hz. We keep the 60 electrodes
that have a standard 10-05 position and are present in every recording, convert to microvolts,
resample to 200\,Hz, remove each window's per-channel median and cut 5\,s windows (4{,}719 in
total). The 149 participants are split into five folds stratified by group, identical for every
model; in each fold the logistic regression's $C\in\{10^{-9},\dots,10\}$ is chosen by AUROC on 20\%
of the training participants, and the classifier is refitted on all training participants. The
diagnosis is made per participant, as the mean probability of their windows; confidence intervals
resample participants.

\subsection{Ablation experiments}
\label{sec:ablations}

\paragraph{Pre-trained against random weights.}
To separate what pre-training contributes from what the architecture and the task head contribute, we
evaluate the same network with random weights through the identical feature extraction, probe and
fine-tuning protocol, on TUEV, TUAB, Siena, Bonn and Mumtaz.

\paragraph{Diffusion against flow matching.}
Because the objective is a single switch in our implementation, we trained a second model,
\model{}-F, that differs from \model{} only in the objective (rectified flow,
\cref{sec:objective}): architecture, data, optimiser and number of training epochs are identical. The two are
compared on every downstream dataset with a paired bootstrap on the shared test samples.

\paragraph{Electrode order and missing electrodes.}
\Cref{sec:properties} shows that \model{} is equivariant to the order of its electrodes by
construction. We test this on the Mumtaz test set (19 electrodes, 1{,}179 windows) by listing the
electrodes in five random orders, each electrode keeping its own position, and comparing every gate
activation and every test prediction with those of the original order. As a control, we shuffle the
positions while leaving the signals in place. To test missing electrodes, the linear probe is trained
once on the full 19-electrode montage and applied, unchanged, to the same test recordings with only
14, 10 or 5 electrodes given, five random subsets for each size; no electrode is imputed.

\subsection{Scaling experiment: NOMAD-EEG-Large}
\label{sec:scaling}

\model{} was pre-trained on 147\,h from one corpus with one montage. To test whether the
architecture improves when it is made larger and given more, and more varied, EEG, we trained
\model{}-Large, which scales the model and the pre-training data together, and evaluated it with
exactly the protocols used for Base.

\paragraph{Model.}
\model{}-Large uses the same code and the same block as Base (\cref{sec:arch}), widened
from $d=256$ to $d=512$, deepened from 16 to 24 blocks and given 8 instead of 4 attention heads:
67.6\,M parameters against 12.0\,M (5.6$\times$). Features are read from blocks 6, 12, 18 and 24
with the same windowed mean and standard deviation, which gives 20{,}480 dimensions instead of
10{,}240 (\cref{tab:scaling_setup}).

\paragraph{Pre-training data.}
The pool contains 6{,}984\,h of EEG from seven corpora (47$\times$ Base): Sleep-EDF expanded
(2{,}954\,h), TUSZ (1{,}202\,h, training and development splits), TUAB (1{,}024\,h, training split),
CHB-MIT (990\,h), TUEP (627\,h), TUEV (147\,h, as for Base) and SEED-V (41\,h). Recordings keep their
native montages, with every channel that has a standard 10-20 or 10-05 position (bipolar
derivations at the midpoint of their two electrodes), which gives 38 distinct channel groups with up
to 62 electrodes. Signals are resampled to 200\,Hz and cut into 30\,s windows, and each window's
per-channel median is removed. The evaluation patients of TUSZ and TUAB and the Sleep-EDF subjects
used downstream were held out of the pool, as were Siena, Bonn, Mumtaz, EEGMMIDB and BCIAUT. TUEP
and SEED-V were included in full, so for Large these two evaluations are in-domain.

\paragraph{Training.}
Each batch is drawn from one channel group, with a random subset of 2 to 32 of its electrodes and a
crop of 30\,s half of the time and 5 to 25\,s otherwise, capped at 640 electrode-seconds per sample.
Corpora are sampled with probability proportional to $\sqrt{\min(\text{hours},1{,}200)}$, so the
largest corpora do not dominate. We count one epoch as as many samples as the pool has 30\,s windows
(about 838{,}000); because samples are crops of at most 30\,s on a subset of electrodes, this is an
upper bound on the signal seen. The optimiser, schedule, moving average, precision and effective batch
of 32 are those of Base (\cref{sec:impl}), with learning rate $2\times10^{-4}$, a warm-up over the
first 0.08 epochs and 2.3 epochs in total (60{,}000 weight updates). Large was trained on a single
20\,GB slice of an A100 GPU (4 samples per update with 8-step gradient accumulation) in 12-hour jobs
that resume from the last checkpoint, and training ran to completion. We evaluate two checkpoints,
after 1.0 and 1.9 epochs. By 1.9 epochs the learning rate had decayed to 5\% of its peak and the probe
results changed little between the two checkpoints on most datasets (\cref{tab:scaling}), so the final
0.3 epochs, at an even lower learning rate, were not evaluated.

\begin{table}[!htbp]
\centering
\caption{Base and Large configurations. An epoch is as many samples as the training set has 30\,s
windows; for Large, whose samples are crops of at most 30\,s, it is an upper bound on the signal seen.}
\label{tab:scaling_setup}
\small
\begin{tabular}{@{}l c c@{}}
\toprule
 & \model{} (Base) & \model{}-Large \\
\midrule
Parameters & 12.0\,M & 67.6\,M \\
Width / blocks / heads & 256 / 16 / 4 & 512 / 24 / 8 \\
Feature dimension & 10{,}240 & 20{,}480 \\
Pre-training data & 147\,h, 1 corpus & 6{,}984\,h, 7 corpora \\
Montages & 1 (21 electrodes) & 38 channel groups, 1--62 electrodes \\
Electrodes per sample & 21 & random subset of 2--32 \\
Training windows & 17{,}325 & $\approx$838{,}000 \\
Effective batch & 32 & 32 \\
Training length & 74 epochs & 2.3 epochs \\
Checkpoint evaluated & after 74 epochs & after 1.9 epochs \\
Samples seen at that checkpoint & 1.28\,M & 1.63\,M \\
\bottomrule
\end{tabular}
\end{table}

\paragraph{Evaluation.}
Large is evaluated after 1.0 and 1.9 epochs on all ten benchmarks, with the frozen linear
probes, splits and metrics of \cref{sec:protocol}; it is not fine-tuned. Its input windows have the
per-channel median removed, as in its pre-training; all other preprocessing is that of Base.

\subsection{Model interpretation}
\label{sec:attribution_methods}

To see what drives \model{}'s seizure decisions, we attribute the seizure score of the frozen model
with its TUSZ linear probe (\cref{tab:tusz}) to the input with Integrated
Gradients~\citep{sundararajan2017ig}: the gradient of the probe's seizure logit with respect to the
input, integrated over 32 steps along a straight path from a baseline in which every channel is at its
median. Positive attributions (evidence for seizure) are summed per electrode over 0.25\,s bins and over
electrodes per bin. Two 5\,s segments are drawn from the TUSZ test set, whose patients the probe never
saw, by a rule fixed in advance: a seeded random choice among test windows labelled seizure that the
model classifies with probability above 0.9, and one among windows labelled non-seizure with
probability below 0.1, so that they are typical correct decisions rather than hand-picked ones.

\subsection{Use of AI tools}
\label{sec:ai}
During the preparation of this work, the authors used Claude to improve language
and readability. After using these tools, the authors reviewed and edited the content as needed and take full responsibility for the content of the published article.

\section{Results}
\label{sec:results}
\subsection{Overview}
\label{sec:overview}

\begin{figure}[!htbp]
\centering
\includegraphics[width=0.72\textwidth]{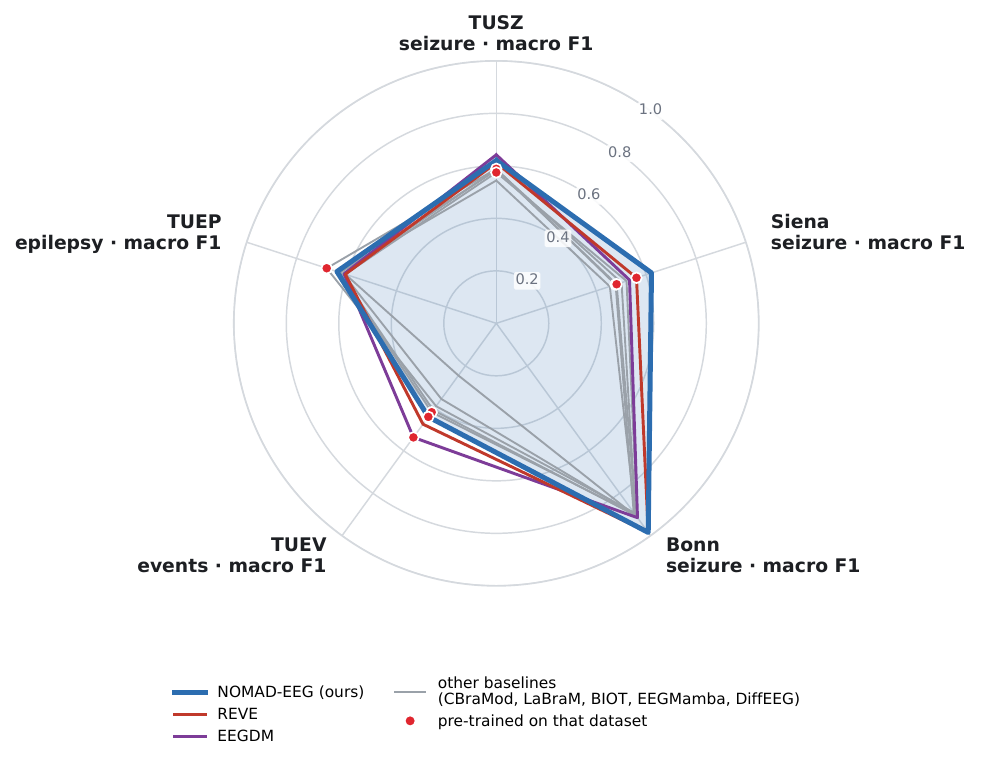}
\caption{\textbf{Summary of the epilepsy tasks.} Macro F1 of the frozen linear probe of every model
on seizure detection (TUSZ, Siena, Bonn), epileptiform event classification (TUEV, mean of the six
per-class F1 values) and epilepsy diagnosis (TUEP, patient level). Macro F1 is shown because it
measures decisions at the classifier's threshold, which in seizure detection determines the false
alarms; the primary metric of each dataset is given in its table. Red dots: the model's
pre-training data contained that dataset (\cref{sec:baselines}); LaBraM's TUEP value is shown for
completeness although it is excluded from \cref{tab:tuep} for that reason. Values come from the same
probes as \cref{tab:tusz,tab:siena,tab:bonn,tab:tuev,tab:tuep}; the TUEV macro F1 is computed from the
per-class F1 values of those predictions.}
\label{fig:radar}
\end{figure}

Across the eight benchmarks compared with all baselines (\cref{tab:tuab,tab:tusz,tab:tuev,tab:siena,tab:bonn,tab:mumtaz,tab:eegmmidb,tab:sleepedf}),
\model{} is never ranked below fourth by each dataset's primary metric (AUROC for binary, balanced
accuracy for multi-class tasks) and is first on Bonn and EEGMMIDB; REVE and EEGDM lead most often,
and \model{} ranks ahead of LaBraM, CBraMod, BIOT, EEGMamba and DiffEEG on average. On the five
epilepsy tasks (\cref{fig:radar}), \model{} has the highest macro F1 on Siena (0.622) and Bonn (0.985,
level with REVE), the highest on TUEP among the models not pre-trained on it (0.638), the second
highest on TUSZ (0.620, behind EEGDM's 0.642), and is behind EEGDM and REVE on TUEV (0.440 against
0.537 and 0.475).

\subsection{TUH benchmarks (TUAB, TUSZ, TUEV, TUEP)}
\label{sec:res_tuh}

On TUEV, \model{} fine-tuned end to end reaches $\kappa$ 0.635 and weighted F1 0.814, the highest
values in \cref{tab:tuev}, with confidence intervals above those of every other model; only on
balanced accuracy is EEGDM's frozen probe higher (0.647 against 0.597). As a frozen probe, \model{} is
third on TUEV (balanced accuracy 0.542, $\kappa$ 0.465). On TUSZ the top three frozen probes are
close: \model{}'s AUROC of 0.866 is 0.013 below REVE's, and its macro F1 (0.620) is second only to
EEGDM's (0.642); fine-tuned, \model{} has the highest macro F1 of all models (0.676), while REVE and
EEGDM keep the highest AUROC, PR-AUC and balanced accuracy (\cref{tab:tusz}). On TUAB, REVE leads on
every metric and \model{} is fourth by AUROC (0.838), behind REVE, EEGDM and BIOT (\cref{tab:tuab}).
Apart from TUSZ's macro F1, fine-tuning does not improve on the frozen probe on these two datasets:
TUSZ AUROC 0.854 [0.847, 0.860] against 0.866, with PR-AUC falling from 0.446 to 0.345 and balanced
accuracy from 0.787 to 0.773; TUAB AUROC 0.833 against 0.838, balanced accuracy 0.770 for both. The
selected fine-tuning epochs reached a validation balanced accuracy of 0.961 (TUSZ) and 0.981 (TUAB).

On TUEP all seven compared models separate the two groups above chance from frozen features
(\cref{tab:tuep}). \model{} has the highest patient-level accuracy (0.640) and F1 (0.638). Only
EEGMamba, pre-trained on the TUH corpus that contains TUEP, has a higher AUROC (0.694 against 0.679)
and PR-AUC, and no model is significantly better than \model{} on either (paired difference with
EEGMamba, 95\% CI $[-0.083, 0.047]$). Among the four models pre-trained neither on TUEP nor on the
full TUH corpus (\model{}, EEGDM, BIOT and DiffEEG), \model{} is best on every metric.

\begin{table}[!htbp]
\centering
\caption{TUAB abnormal-EEG detection (5\,s windows; 50{,}000 test windows), frozen linear probe. Brackets: 95\% bootstrap CI. TUAB's training split was in BIOT's pre-training data, and CBraMod and EEGMamba were pre-trained on the TUH corpus it is drawn from (\cref{sec:baselines}). The indented row is \model{} fine-tuned end to end (not part of the bold comparison); its epoch was selected on the benchmark's validation windows, which come from the training patients (\cref{sec:finetune}).}
\label{tab:tuab}
\small
\begin{tabular}{@{}l r c c c c@{}}
\toprule
Model & Dim & AUROC & PR-AUC & Bal.\ acc. & Macro F1 \\
\midrule
\model{} (ours) & 10{,}240 & 0.838 {\scriptsize[0.834, 0.842]} & 0.820 & 0.770 & 0.771 \\
\quad fine-tuned & --- & 0.833 {\scriptsize[0.830, 0.837]} & 0.785 & 0.770 & 0.771 \\
\midrule
EEGDM~\citep{puah2025eegdm} & 2{,}560 & 0.869 {\scriptsize[0.865, 0.872]} & 0.862 & 0.792 & 0.792 \\
REVE~\citep{ouahidi2025reve} & 512 & \textbf{0.870} {\scriptsize[0.867, 0.873]} & \textbf{0.880} & \textbf{0.800} & \textbf{0.802} \\
CBraMod~\citep{wang2025cbramod} & 200 & 0.808 {\scriptsize[0.804, 0.812]} & 0.803 & 0.738 & 0.739 \\
LaBraM~\citep{jiang2024labram} & 200 & 0.788 {\scriptsize[0.784, 0.792]} & 0.785 & 0.713 & 0.714 \\
BIOT~\citep{yang2023biot} & 256 & 0.847 {\scriptsize[0.844, 0.850]} & 0.849 & 0.771 & 0.772 \\
EEGMamba~\citep{wang2025eegmamba} & 200 & 0.746 {\scriptsize[0.742, 0.750]} & 0.726 & 0.689 & 0.690 \\
DiffEEG~\citep{diffeeg2026} & 2{,}400 & 0.760 {\scriptsize[0.756, 0.764]} & 0.718 & 0.695 & 0.696 \\
\bottomrule
\end{tabular}
\end{table}

\begin{table}[!htbp]
\centering
\caption{TUSZ seizure detection (5\,s windows; 50{,}000 test windows at the true seizure prevalence), frozen linear probe. TUSZ was in LaBraM's and DiffEEG's pre-training data, and REVE, CBraMod and EEGMamba were pre-trained on the TUH corpus it is drawn from (\cref{sec:baselines}). The indented row is \model{} fine-tuned end to end (not part of the bold comparison); its epoch was selected on the benchmark's validation windows, which come from the training patients (\cref{sec:finetune}).}
\label{tab:tusz}
\small
\begin{tabular}{@{}l r c c c c@{}}
\toprule
Model & Dim & AUROC & PR-AUC & Bal.\ acc. & Macro F1 \\
\midrule
\model{} (ours) & 10{,}240 & 0.866 {\scriptsize[0.860, 0.873]} & 0.446 & 0.787 & 0.620 \\
\quad fine-tuned & --- & 0.854 {\scriptsize[0.847, 0.860]} & 0.345 & 0.773 & 0.676 \\
\midrule
EEGDM~\citep{puah2025eegdm} & 2{,}560 & 0.874 {\scriptsize[0.866, 0.881]} & \textbf{0.527} & \textbf{0.801} & \textbf{0.642} \\
REVE~\citep{ouahidi2025reve} & 512 & \textbf{0.879} {\scriptsize[0.873, 0.886]} & 0.463 & 0.798 & 0.611 \\
CBraMod~\citep{wang2025cbramod} & 200 & 0.855 {\scriptsize[0.847, 0.862]} & 0.455 & 0.776 & 0.586 \\
LaBraM~\citep{jiang2024labram} & 200 & 0.845 {\scriptsize[0.837, 0.851]} & 0.416 & 0.765 & 0.589 \\
BIOT~\citep{yang2023biot} & 256 & 0.760 {\scriptsize[0.752, 0.768]} & 0.279 & 0.691 & 0.544 \\
EEGMamba~\citep{wang2025eegmamba} & 200 & 0.836 {\scriptsize[0.829, 0.843]} & 0.308 & 0.763 & 0.585 \\
DiffEEG~\citep{diffeeg2026} & 2{,}400 & 0.813 {\scriptsize[0.804, 0.820]} & 0.301 & 0.745 & 0.575 \\
\bottomrule
\end{tabular}
\end{table}

\begin{table}[!htbp]
\centering
\caption{TUEV event classification (6 classes; 29{,}421 test events; in the pre-training data of \model{}, EEGDM and BIOT, and in the TUH corpus on which CBraMod and EEGMamba were pre-trained, \cref{sec:baselines}), frozen linear probe with 5-fold subject-grouped cross-validation. The indented row is \model{} fine-tuned end to end; bold marks the best value in each column including this row, although the baselines were evaluated only as frozen probes.}
\label{tab:tuev}
\small
\begin{tabular}{@{}l r c c c@{}}
\toprule
Model & Dim & Bal.\ acc. & Cohen $\kappa$ & Weighted F1 \\
\midrule
\model{} (ours) & 10{,}240 & 0.542 {\scriptsize[0.529, 0.554]} & 0.465 & 0.719 \\
\quad fine-tuned & --- & 0.597 {\scriptsize[0.588, 0.605]} & \textbf{0.635} & \textbf{0.814} \\
\midrule
EEGDM~\citep{puah2025eegdm} & 2{,}560 & \textbf{0.647} {\scriptsize[0.638, 0.657]} & 0.568 & 0.783 \\
REVE~\citep{ouahidi2025reve} & 512 & 0.565 {\scriptsize[0.553, 0.577]} & 0.515 & 0.753 \\
CBraMod~\citep{wang2025cbramod} & 200 & 0.466 {\scriptsize[0.455, 0.477]} & 0.346 & 0.635 \\
LaBraM~\citep{jiang2024labram} & 200 & 0.521 {\scriptsize[0.510, 0.534]} & 0.365 & 0.633 \\
BIOT~\citep{yang2023biot} & 256 & 0.519 {\scriptsize[0.507, 0.530]} & 0.428 & 0.698 \\
EEGMamba~\citep{wang2025eegmamba} & 200 & 0.520 {\scriptsize[0.509, 0.530]} & 0.374 & 0.664 \\
DiffEEG~\citep{diffeeg2026} & 2{,}400 & 0.329 {\scriptsize[0.318, 0.340]} & 0.179 & 0.503 \\
\bottomrule
\end{tabular}
\end{table}

\begin{table}[!htbp]
\centering
\caption{TUEP epilepsy diagnosis (200 patients, 100 with and 100 without epilepsy; 19 electrodes, 60 random 5\,s windows per patient), frozen encoders and a linear probe evaluated on unseen patients (5-fold cross-validation stratified by diagnosis, identical folds for every model). A patient's score is the mean probability of their windows. Brackets: 95\% CI from 1{,}000 bootstrap resamples of patients; accuracy and F1 at a 0.5 threshold. LaBraM is not included because TUEP was part of its pre-training data (under the same protocol its patient-level AUROC is 0.741 and accuracy 0.680). $^\ddagger$The model was pre-trained on the TUH corpus from which TUEP is drawn (\cref{sec:baselines}). Bold: best among the models whose pre-training did not include TUEP recordings (no mark); underlined: best overall.}
\label{tab:tuep}
\small
\begin{tabular}{@{}l r c c c c c@{}}
\toprule
Model & Dim & AUROC & PR-AUC & Accuracy & Weighted F1 & Macro F1 \\
\midrule
\model{} (ours) & 10{,}240 & \textbf{0.679} {\scriptsize[0.607, 0.751]} & \textbf{0.675} & \underline{\textbf{0.640}} & \underline{\textbf{0.638}} & \underline{\textbf{0.638}} \\
EEGDM~\citep{puah2025eegdm} & 2{,}560 & 0.673 {\scriptsize[0.599, 0.745]} & 0.638 & 0.610 & 0.609 & 0.609 \\
REVE~\citep{ouahidi2025reve}$^\ddagger$ & 512 & 0.671 {\scriptsize[0.596, 0.748]} & 0.674 & 0.605 & 0.605 & 0.605 \\
CBraMod~\citep{wang2025cbramod}$^\ddagger$ & 200 & 0.671 {\scriptsize[0.591, 0.745]} & 0.667 & 0.615 & 0.615 & 0.615 \\
BIOT~\citep{yang2023biot} & 256 & 0.675 {\scriptsize[0.605, 0.748]} & 0.633 & 0.630 & 0.630 & 0.630 \\
EEGMamba~\citep{wang2025eegmamba}$^\ddagger$ & 200 & \underline{0.694} {\scriptsize[0.619, 0.766]} & \underline{0.701} & 0.625 & 0.625 & 0.625 \\
DiffEEG~\citep{diffeeg2026} & 2{,}400 & 0.622 {\scriptsize[0.542, 0.702]} & 0.603 & 0.605 & 0.605 & 0.605 \\
\bottomrule
\end{tabular}
\end{table}

\subsection{Seizure detection on external corpora (Siena, Bonn)}
\label{sec:res_external}

On Siena, where seizures make up 0.6\% of training windows, \model{} has the highest macro F1 (0.622,
with a CI that excludes every other model), accuracy (0.979) and weighted F1 (0.984). EEGDM ranks
seizures best by AUROC (0.915) and REVE has the highest PR-AUC (\cref{tab:siena}). Bonn is close to
saturated: \model{} and REVE both reach an AUROC of about 0.998--0.999 and all confidence intervals
overlap (\cref{tab:bonn}). Fed the single signal-carrying electrode alone, rather than a 22-channel
frame, \model{} still reaches an AUROC of 0.995.

\begin{table}[!htbp]
\centering
\caption{Siena seizure detection (external to \model{}; 18{,}729 test windows, 0.6\% seizure in training), frozen linear probe. Siena recordings were in REVE's, LaBraM's and EEGMamba's pre-training data (\cref{sec:baselines}).}
\label{tab:siena}
\small
\begin{tabular}{@{}l r c c c c c@{}}
\toprule
Model & Dim & AUROC & PR-AUC & Accuracy & Weighted F1 & Macro F1 \\
\midrule
\model{} (ours) & 10{,}240 & 0.882 {\scriptsize[0.847, 0.914]} & 0.306 & \textbf{0.979} & \textbf{0.984} & \textbf{0.622} \\
EEGDM~\citep{puah2025eegdm} & 2{,}560 & \textbf{0.915} {\scriptsize[0.889, 0.938]} & 0.248 & 0.916 & 0.950 & 0.534 \\
REVE~\citep{ouahidi2025reve} & 512 & 0.906 {\scriptsize[0.872, 0.937]} & \textbf{0.460} & 0.946 & 0.966 & 0.562 \\
CBraMod~\citep{wang2025cbramod} & 200 & 0.850 {\scriptsize[0.806, 0.888]} & 0.189 & 0.872 & 0.925 & 0.502 \\
LaBraM~\citep{jiang2024labram} & 200 & 0.797 {\scriptsize[0.756, 0.836]} & 0.056 & 0.833 & 0.903 & 0.478 \\
BIOT~\citep{yang2023biot} & 256 & 0.707 {\scriptsize[0.657, 0.755]} & 0.020 & 0.780 & 0.870 & 0.455 \\
EEGMamba~\citep{wang2025eegmamba} & 200 & 0.855 {\scriptsize[0.812, 0.893]} & 0.252 & 0.828 & 0.899 & 0.482 \\
DiffEEG~\citep{diffeeg2026} & 2{,}400 & 0.777 {\scriptsize[0.724, 0.827]} & 0.057 & 0.921 & 0.952 & 0.522 \\
\bottomrule
\end{tabular}
\end{table}

\begin{table}[!htbp]
\centering
\caption{Bonn seizure detection (external; single-channel recordings; 100 test segments), frozen linear probe. All models read the recording in a 22-channel frame; the indented row feeds \model{} the one signal-carrying electrode alone.}
\label{tab:bonn}
\small
\begin{tabular}{@{}l r c c c c@{}}
\toprule
Model & Dim & AUROC & PR-AUC & Bal.\ acc. & Macro F1 \\
\midrule
\model{} (ours) & 10{,}240 & \textbf{0.999} {\scriptsize[0.996, 1.000]} & \textbf{0.998} & \textbf{0.994} & \textbf{0.985} \\
\quad single electrode only & 10{,}240 & 0.995 {\scriptsize[0.981, 1.000]} & 0.983 & 0.944 & 0.913 \\
\midrule
EEGDM~\citep{puah2025eegdm} & 2{,}560 & 0.994 {\scriptsize[0.982, 1.000]} & 0.975 & 0.963 & 0.915 \\
REVE~\citep{ouahidi2025reve} & 512 & 0.998 {\scriptsize[0.991, 1.000]} & 0.992 & \textbf{0.994} & \textbf{0.985} \\
CBraMod~\citep{wang2025cbramod} & 200 & 0.996 {\scriptsize[0.986, 1.000]} & 0.985 & 0.956 & 0.903 \\
LaBraM~\citep{jiang2024labram} & 200 & 0.995 {\scriptsize[0.983, 1.000]} & 0.982 & 0.950 & 0.890 \\
BIOT~\citep{yang2023biot} & 256 & 0.979 {\scriptsize[0.943, 1.000]} & 0.949 & 0.938 & 0.900 \\
EEGMamba~\citep{wang2025eegmamba} & 200 & 0.948 {\scriptsize[0.867, 1.000]} & 0.919 & 0.906 & 0.906 \\
DiffEEG~\citep{diffeeg2026} & 2{,}400 & 0.982 {\scriptsize[0.954, 0.999]} & 0.937 & 0.944 & 0.913 \\
\bottomrule
\end{tabular}
\end{table}

\subsection{New montages and tasks (Mumtaz, EEGMMIDB, Sleep-EDF, SEED-V)}
\label{sec:res_new}

On EEGMMIDB, the 64-electrode motor-imagery benchmark in which 45 of the electrode positions were
never seen in \model{}'s pre-training, \model{} is the best frozen model on all three metrics
(balanced accuracy 0.442, chance 0.25; \cref{tab:eegmmidb}); the second-best model, DiffEEG (0.418),
has an overlapping CI, and \model{}'s CI excludes REVE (0.371) and every
model below it. On Sleep-EDF, with two bipolar channels and 30\,s epochs, \model{} is third by
balanced accuracy (0.757), level with REVE (0.760) and behind EEGDM (0.779), and second by Cohen's
$\kappa$ (\cref{tab:sleepedf}). On Mumtaz \model{} is third by AUROC (0.942) behind REVE and EEGDM,
with overlapping CIs (\cref{tab:mumtaz}). Fine-tuning does not improve on the frozen probe on Mumtaz
(0.821 against 0.832 balanced accuracy) or Sleep-EDF (0.758 against 0.757), whereas on EEGMMIDB it
does (0.596 against 0.442). On SEED-V, five-class emotion recognition from 59 electrodes in 1\,s
windows evaluated for \model{} alone, the frozen probe reaches a balanced accuracy of 0.315 (chance
0.20) and $\kappa$ 0.136, and fine-tuning raises them to 0.384 and 0.231 (\cref{tab:seedv}).

\begin{table}[!htbp]
\centering
\caption{Mumtaz depression vs.\ healthy (external; 19 electrodes, 5\,s windows; 1{,}179 test windows from 10 held-out subjects), frozen linear probe. The indented row is \model{} fine-tuned end to end (no baseline was fine-tuned).}
\label{tab:mumtaz}
\small
\begin{tabular}{@{}l r c c c c@{}}
\toprule
Model & Dim & AUROC & PR-AUC & Bal.\ acc. & Macro F1 \\
\midrule
\model{} (ours) & 10{,}240 & 0.942 {\scriptsize[0.929, 0.954]} & 0.951 & 0.832 & 0.831 \\
\quad fine-tuned & --- & 0.872 {\scriptsize[0.851, 0.893]} & 0.792 & 0.821 & 0.821 \\
\midrule
EEGDM~\citep{puah2025eegdm} & 2{,}560 & 0.946 {\scriptsize[0.935, 0.958]} & 0.945 & \textbf{0.857} & \textbf{0.856} \\
REVE~\citep{ouahidi2025reve} & 512 & \textbf{0.960} {\scriptsize[0.951, 0.968]} & \textbf{0.962} & 0.842 & 0.841 \\
CBraMod~\citep{wang2025cbramod} & 200 & 0.860 {\scriptsize[0.838, 0.883]} & 0.810 & 0.804 & 0.803 \\
LaBraM~\citep{jiang2024labram} & 200 & 0.904 {\scriptsize[0.887, 0.922]} & 0.886 & 0.838 & 0.837 \\
BIOT~\citep{yang2023biot} & 256 & 0.928 {\scriptsize[0.912, 0.942]} & 0.910 & 0.829 & 0.829 \\
EEGMamba~\citep{wang2025eegmamba} & 200 & 0.867 {\scriptsize[0.845, 0.887]} & 0.843 & 0.810 & 0.809 \\
DiffEEG~\citep{diffeeg2026} & 2{,}400 & 0.901 {\scriptsize[0.883, 0.918]} & 0.889 & 0.807 & 0.806 \\
\bottomrule
\end{tabular}
\end{table}

\begin{table}[!htbp]
\centering
\caption{EEGMMIDB motor imagery (external to \model{}; 4 classes, 64 electrodes, 4\,s trials; 1{,}803 test trials from 20 held-out subjects), frozen linear probe. EEGMMIDB was in LaBraM's pre-training data (\cref{sec:baselines}). The indented row is \model{} fine-tuned end to end.}
\label{tab:eegmmidb}
\small
\begin{tabular}{@{}l r c c c@{}}
\toprule
Model & Dim & Bal.\ acc. & Cohen $\kappa$ & Weighted F1 \\
\midrule
\model{} (ours) & 10{,}240 & \textbf{0.442} {\scriptsize[0.419, 0.463]} & \textbf{0.256} & \textbf{0.442} \\
\quad fine-tuned & --- & 0.596 {\scriptsize[0.574, 0.618]} & 0.461 & 0.599 \\
\midrule
EEGDM~\citep{puah2025eegdm} & 2{,}560 & 0.347 {\scriptsize[0.325, 0.367]} & 0.130 & 0.338 \\
REVE~\citep{ouahidi2025reve} & 512 & 0.371 {\scriptsize[0.348, 0.393]} & 0.161 & 0.370 \\
CBraMod~\citep{wang2025cbramod} & 200 & 0.308 {\scriptsize[0.287, 0.330]} & 0.078 & 0.306 \\
LaBraM~\citep{jiang2024labram} & 200 & 0.302 {\scriptsize[0.280, 0.324]} & 0.070 & 0.301 \\
BIOT~\citep{yang2023biot} & 256 & 0.274 {\scriptsize[0.255, 0.294]} & 0.032 & 0.269 \\
EEGMamba~\citep{wang2025eegmamba} & 200 & 0.304 {\scriptsize[0.283, 0.324]} & 0.073 & 0.303 \\
DiffEEG~\citep{diffeeg2026} & 2{,}400 & 0.418 {\scriptsize[0.395, 0.439]} & 0.224 & 0.417 \\
\bottomrule
\end{tabular}
\end{table}

\begin{table}[!htbp]
\centering
\caption{Sleep-EDF-20 sleep staging (external; 5 stages, 2 bipolar channels, 30\,s epochs; 9{,}137 test epochs from 4 held-out subjects), frozen linear probe. The indented row is \model{} fine-tuned end to end.}
\label{tab:sleepedf}
\small
\begin{tabular}{@{}l r c c c@{}}
\toprule
Model & Dim & Bal.\ acc. & Cohen $\kappa$ & Weighted F1 \\
\midrule
\model{} (ours) & 10{,}240 & 0.757 {\scriptsize[0.748, 0.766]} & 0.680 & 0.770 \\
\quad fine-tuned & --- & 0.758 {\scriptsize[0.748, 0.768]} & 0.694 & 0.784 \\
\midrule
EEGDM~\citep{puah2025eegdm} & 2{,}560 & \textbf{0.779} {\scriptsize[0.770, 0.788]} & \textbf{0.718} & \textbf{0.802} \\
REVE~\citep{ouahidi2025reve} & 512 & 0.760 {\scriptsize[0.750, 0.770]} & 0.675 & 0.771 \\
CBraMod~\citep{wang2025cbramod} & 200 & 0.725 {\scriptsize[0.715, 0.735]} & 0.639 & 0.743 \\
LaBraM~\citep{jiang2024labram} & 200 & 0.738 {\scriptsize[0.729, 0.747]} & 0.676 & 0.770 \\
BIOT~\citep{yang2023biot} & 256 & 0.741 {\scriptsize[0.731, 0.750]} & 0.659 & 0.761 \\
EEGMamba~\citep{wang2025eegmamba} & 200 & 0.722 {\scriptsize[0.712, 0.732]} & 0.601 & 0.717 \\
DiffEEG~\citep{diffeeg2026} & 2{,}400 & 0.715 {\scriptsize[0.704, 0.725]} & 0.628 & 0.737 \\
\bottomrule
\end{tabular}
\end{table}

\begin{table}[!htbp]
\centering
\caption{SEED-V emotion recognition (external; 5 emotions, 59 electrodes, 1\,s windows; within every session film clips 1--5 train, 6--10 validation, 11--15 test, all 16 subjects in every split), \model{} only. Chance balanced accuracy is 0.20.}
\label{tab:seedv}
\small
\begin{tabular}{@{}l c c c@{}}
\toprule
\model{} & Bal.\ acc. & Cohen $\kappa$ & Weighted F1 \\
\midrule
frozen linear probe & 0.315 {\scriptsize[0.310, 0.319]} & 0.136 & 0.302 \\
fine-tuned & 0.384 {\scriptsize[0.380, 0.389]} & 0.231 & 0.392 \\
\bottomrule
\end{tabular}
\end{table}

\subsection{Transfer to unseen populations}
\label{sec:res_transfer}

\model{} is the best of the five models on every BCIAUT-P300 metric (\cref{tab:bciaut}). It reaches an
AUROC of 0.718 and selects the attended object in 75.5\% of blocks after ten runs, against 61.2\% for
EEGDM and 57.5\% for REVE. The advantage holds within subjects: its per-subject AUROC exceeds REVE's in
all 15 subjects and EEGDM's in 13. BIOT and DiffEEG are close to chance.

\begin{table}[t]
\centering
\caption{P300 detection on BCIAUT-P300 (15 participants with ASD, 168{,}000 epochs,
12.5\% targets). Encoders are frozen and never trained on this corpus; a linear probe is
evaluated on subjects it never saw (five folds of three subjects). Brackets: 95\% CI from a
bootstrap over subjects. ``Object acc.'' is the rate at which the attended object among eight is
selected from the first 1, 3 or 10 runs of a block (chance 12.5\%).}
\label{tab:bciaut}
\footnotesize
\setlength{\tabcolsep}{3.5pt}
\resizebox{\textwidth}{!}{%
\begin{tabular}{@{}l c l c c c c@{}}
\toprule
Model & Channels read & AUROC & Bal.\ acc. & PR-AUC & Object acc.\ (1/3/10 runs) & Per-subject AUROC range \\
\midrule
\textbf{\model{}} & 8 & \textbf{0.718} [0.695, 0.739] & \textbf{0.660} & \textbf{0.289} & \textbf{32.6 / 51.5 / 75.5\%} & 0.652--0.805 \\
EEGDM & 4 bipolar & 0.667 [0.631, 0.699] & 0.624 & 0.227 & 27.8 / 42.4 / 61.2\% & 0.540--0.805 \\
REVE & 8 & 0.651 [0.626, 0.675] & 0.612 & 0.209 & 24.6 / 37.9 / 57.5\% & 0.563--0.752 \\
BIOT & 2 bipolar & 0.530 [0.521, 0.539] & 0.520 & 0.136 & 14.5 / 17.2 / 21.8\% & 0.503--0.571 \\
DiffEEG & 6 & 0.522 [0.508, 0.534] & 0.517 & 0.133 & 13.9 / 15.1 / 19.4\% & 0.457--0.556 \\
\bottomrule
\end{tabular}}
\end{table}

All five models detect Parkinson's disease above chance from frozen features, but here \model{} is
not the best (\cref{tab:parkinson}). EEGDM reaches a participant-level AUROC of 0.795
[0.714, 0.868], significantly above \model{}'s 0.684 [0.590, 0.777] (paired difference
$-0.111$, 95\% CI $[-0.203, -0.012]$); \model{} is level with BIOT, DiffEEG and REVE, whose
intervals overlap its own. \model{}'s selected $C$ ranged from $10^{-4}$ to 10 across folds.

\begin{table}[!htbp]
\centering
\caption{Parkinson's disease vs.\ control (OpenNeuro ds004584; 149 participants, 100 PD / 49 controls; 60 electrodes, 5\,s windows), frozen encoders and a linear probe evaluated on unseen participants (5-fold cross-validation stratified by group, identical folds for every model). A participant's score is the mean probability of their windows. Brackets: 95\% CI from 1{,}000 bootstrap resamples of participants.}
\label{tab:parkinson}
\small
\begin{tabular}{@{}l c c c@{}}
\toprule
Model & Window AUROC & Participant AUROC & Participant bal.\ acc. \\
\midrule
\model{} (ours) & 0.679 {\scriptsize[0.609, 0.754]} & 0.684 {\scriptsize[0.590, 0.777]} & 0.620 {\scriptsize[0.542, 0.706]} \\
EEGDM~\citep{puah2025eegdm} & \textbf{0.783} {\scriptsize[0.709, 0.845]} & \textbf{0.795} {\scriptsize[0.714, 0.868]} & \textbf{0.702} {\scriptsize[0.618, 0.782]} \\
REVE~\citep{ouahidi2025reve} & 0.653 {\scriptsize[0.571, 0.733]} & 0.643 {\scriptsize[0.547, 0.740]} & 0.626 {\scriptsize[0.543, 0.705]} \\
BIOT~\citep{yang2023biot} & 0.659 {\scriptsize[0.583, 0.730]} & 0.681 {\scriptsize[0.593, 0.771]} & 0.672 {\scriptsize[0.593, 0.748]} \\
DiffEEG~\citep{diffeeg2026} & 0.634 {\scriptsize[0.585, 0.683]} & 0.713 {\scriptsize[0.623, 0.794]} & 0.672 {\scriptsize[0.587, 0.751]} \\
\bottomrule
\end{tabular}
\end{table}

\subsection{Ablations}
\label{sec:res_ablations}

\paragraph{Pre-trained against random weights.}
Pre-training improves balanced accuracy on TUEV by 0.107 as a frozen probe and by 0.132 after
fine-tuning, improves TUAB AUROC by 0.040, and improves Siena macro F1 by 0.071, all with
non-overlapping confidence intervals (\cref{tab:pretrain}). On Siena the gain is in precision at the
decision threshold (macro F1; PR-AUC is also higher, but its CIs overlap) and not in ranking: AUROC is
unchanged. On Mumtaz, fine-tuning from random weights matches fine-tuning from the pre-trained model,
and Bonn is saturated for both.

\begin{table}[!htbp]
\centering
\caption{Effect of pre-training: \model{} with its pre-trained weights versus the same architecture with random weights, under the identical feature extraction and protocol. Brackets: 95\% bootstrap CI. $\Delta$ in bold where the two CIs do not overlap.}
\label{tab:pretrain}
\small
\begin{tabular}{@{}l l l c c r@{}}
\toprule
Dataset & Setting & Metric & Pre-trained & Random init. & $\Delta$ \\
\midrule
TUEV & linear probe & Bal.\ acc. & 0.542 {\scriptsize[0.529, 0.554]} & 0.434 {\scriptsize[0.422, 0.446]} & \textbf{+0.107} \\
TUEV & fine-tuned & Bal.\ acc. & 0.597 {\scriptsize[0.588, 0.605]} & 0.465 {\scriptsize[0.455, 0.474]} & \textbf{+0.132} \\
TUAB & linear probe & AUROC & 0.838 {\scriptsize[0.834, 0.842]} & 0.798 {\scriptsize[0.794, 0.801]} & \textbf{+0.040} \\
Siena & linear probe & AUROC & 0.882 {\scriptsize[0.847, 0.914]} & 0.892 {\scriptsize[0.858, 0.924]} & -0.010 \\
Siena & linear probe & Macro F1 & 0.622 {\scriptsize[0.596, 0.646]} & 0.551 {\scriptsize[0.538, 0.563]} & \textbf{+0.071} \\
Siena & linear probe & PR-AUC & 0.306 {\scriptsize[0.221, 0.392]} & 0.192 {\scriptsize[0.134, 0.262]} & +0.114 \\
Bonn & linear probe & AUROC & 0.999 {\scriptsize[0.996, 1.000]} & 0.997 {\scriptsize[0.987, 1.000]} & +0.003 \\
Mumtaz & fine-tuned & Bal.\ acc. & 0.821 {\scriptsize[0.798, 0.844]} & 0.828 {\scriptsize[0.807, 0.851]} & -0.007 \\
\bottomrule
\end{tabular}
\end{table}

\paragraph{Diffusion against flow matching.}
Diffusion is better on four datasets (TUEV, TUAB, Siena macro F1 and Sleep-EDF), flow is better on one
(TUSZ balanced accuracy, by 0.011), and the remaining comparisons show no difference
(\cref{tab:objective}). The largest gaps favour diffusion: 0.086 macro F1 on Siena, 0.055 balanced
accuracy on TUEV and 0.041 on
Sleep-EDF.

\begin{table}[!htbp]
\centering
\caption{Diffusion versus flow matching. Both models share architecture, data, number of steps, optimiser and feature extraction; only the objective differs. Frozen linear probe on each dataset; ``Flow $-$ diffusion'' is the paired difference on the same test samples with a 95\% CI from 2{,}000 paired bootstrap resamples.}
\label{tab:objective}
\small
\begin{tabular}{@{}l l c c r l@{}}
\toprule
Dataset & Metric & Diffusion & Flow & Flow $-$ diffusion  \\
\midrule
TUEV & Bal.\ acc. & 0.542 & 0.486 & -0.055 {\scriptsize[-0.067, -0.044]}  \\
TUAB & AUROC & 0.838 & 0.827 & -0.011 {\scriptsize[-0.013, -0.008]} \\
TUAB & Bal.\ acc. & 0.770 & 0.762 & -0.008 {\scriptsize[-0.011, -0.004]}  \\
TUSZ & AUROC & 0.866 & 0.870 & +0.004 {\scriptsize[-0.002, 0.010]} \\
TUSZ & Bal.\ acc. & 0.787 & 0.798 & +0.011 {\scriptsize[0.003, 0.018]} \\
Siena & AUROC & 0.882 & 0.866 & -0.016 {\scriptsize[-0.037, 0.004]}  \\
Siena & Macro F1 & 0.622 & 0.536 & -0.086 {\scriptsize[-0.103, -0.068]} \\
Bonn & AUROC & 0.999 & 0.996 & -0.003 {\scriptsize[-0.010, 0.000]} \\
Mumtaz & Bal.\ acc. & 0.832 & 0.830 & -0.002 {\scriptsize[-0.007, 0.003]}  \\
EEGMMIDB & Bal.\ acc. & 0.442 & 0.422 & -0.020 {\scriptsize[-0.045, 0.006]}  \\
Sleep-EDF & Bal.\ acc. & 0.757 & 0.716 & -0.041 {\scriptsize[-0.050, -0.033]} \\
\bottomrule
\end{tabular}
\end{table}

\begin{figure}[!tp]
  \centering
  \includegraphics[width=\textwidth]{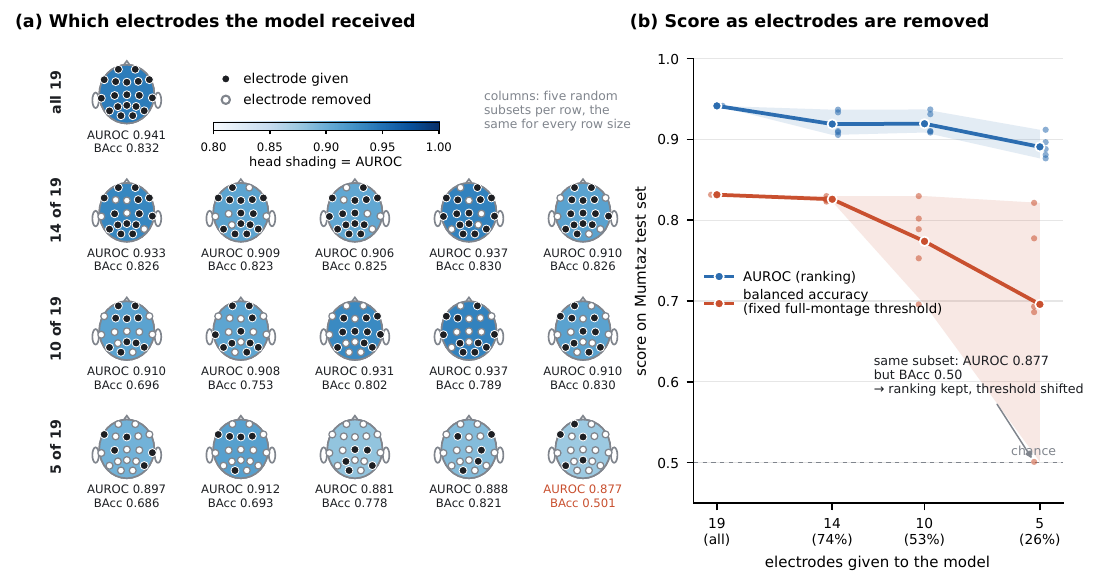}
  \caption{\textbf{Missing-electrode test on Mumtaz} (1{,}179 test windows). The linear probe is
  trained once on the full montage and applied unchanged. (a) Each scalp shows one test condition:
  filled dots are the electrodes given to the model, hollow dots those removed; the shading and the
  label give the AUROC, and BAcc is the balanced accuracy at the probe's threshold. Rows: 19, 14,
  10 and 5 electrodes; columns: five random subsets per size. (b) The same results against the
  number of electrodes given: dots are individual subsets, lines the mean, bands the range over
  subsets. AUROC stays high as electrodes are removed, while balanced accuracy drops because the
  scores shift relative to the threshold fitted on the full montage.}
  \label{fig:missing}
\end{figure}

\paragraph{Electrode order and missing electrodes.}
With the Mumtaz electrodes listed in five random orders, every gate activation equals the original
with its rows permuted (largest relative difference $5.6\times10^{-6}$, floating-point rounding), and
no test prediction changes. Shuffling the positions while leaving the signals in place changes 17.5\%
of the predictions and lowers AUROC from 0.942 to 0.909. With electrodes removed
(\cref{fig:missing}), the mean AUROC falls from 0.942 with all electrodes to 0.919 with 14, 0.919 with
10 and 0.891 with 5, and no subset falls below 0.877. Balanced accuracy at the probe's fixed threshold
degrades more: it stays at 0.826 on average with 14 electrodes but falls to 0.774 with 10 and 0.696
with 5, and to chance (0.501) for one five-electrode subset (Fp1, F7, T5, Fz, Pz) whose AUROC is still
0.877.

\subsection{Scaling model size and pre-training data}
\label{sec:res_scaling}

\Cref{tab:scaling} compares the two models on all ten benchmarks with each dataset's primary metric.
After 1.9 epochs, no Large interval lies above Base's, so \model{}-Large is not significantly better
than Base on any dataset; its point estimate is higher on Bonn and TUEP (below). Its interval lies entirely below Base's on EEGMMIDB (balanced accuracy 0.361 against 0.442),
Sleep-EDF (0.724 against 0.757) and SEED-V (0.301 against 0.315). It is lower, with intervals that
overlap or just touch, on TUSZ (AUROC 0.856 against 0.866), Siena (0.867 against 0.882) and Mumtaz
(0.913 against 0.942). It is
level on Bonn (1.000 against 0.999), TUAB (0.837 against 0.838), TUEV (balanced accuracy 0.542 against
0.542) and TUEP (patient AUROC 0.694 against 0.679; paired difference 95\% CI $[-0.061, 0.031]$).
The highest Large value relative to Base is TUSZ after 1.0 epoch (AUROC 0.878 against
0.866, overlapping intervals), which fell below Base by 1.9 epochs. Between the two checkpoints
Large improved on TUEV (0.450 to 0.542), EEGMMIDB (0.326 to 0.361), TUEP (0.658 to 0.694) and TUAB
(0.828 to 0.837), and changed little elsewhere.

\begin{table}[!htbp]
\centering
\caption{Scaling model size and pre-training data together: \model{} (Base, 12.0\,M parameters, 147\,h of TUEV) against \model{}-Large (67.6\,M parameters, 6{,}984\,h from seven corpora) at two points of its pre-training, under the identical frozen linear probe and splits used for every other table. Each dataset's primary metric with its 95\% bootstrap CI; bold marks the best value in each row. $\downarrow$/$\uparrow$: the Large CI lies entirely below/above Base's (non-overlapping intervals, not a paired test). \emph{In Large pre-training}: whether that dataset's recordings were in Large's unlabelled pre-training data. $^{p}$ patient level. $^{b}$ TUEV is also Base's whole pre-training corpus.}
\label{tab:scaling}
\small
\setlength{\tabcolsep}{4pt}
\begin{tabular}{@{}l l l c c c@{}}
\toprule
Dataset & Metric & In Large & Base & Large & Large \\
 & & pre-training & & step 27{,}000 & step 51{,}000 \\
\midrule
Bonn & AUROC & no & 0.999 {\scriptsize[0.996, 1.000]} & 0.999 {\scriptsize[0.994, 1.000]} & \textbf{1.000} {\scriptsize[1.000, 1.000]} \\
TUSZ & AUROC & training patients & 0.866 {\scriptsize[0.860, 0.873]} & \textbf{0.878} {\scriptsize[0.872, 0.884]} & 0.856 {\scriptsize[0.849, 0.862]} \\
TUAB & AUROC & training patients & \textbf{0.838} {\scriptsize[0.834, 0.842]} & 0.828$\downarrow$ {\scriptsize[0.824, 0.831]} & \textbf{0.837} {\scriptsize[0.834, 0.841]} \\
Siena & AUROC & no & \textbf{0.882} {\scriptsize[0.847, 0.914]} & 0.867 {\scriptsize[0.822, 0.906]} & 0.867 {\scriptsize[0.823, 0.907]} \\
Mumtaz & AUROC & no & \textbf{0.942} {\scriptsize[0.929, 0.954]} & 0.913$\downarrow$ {\scriptsize[0.895, 0.929]} & 0.913 {\scriptsize[0.894, 0.929]} \\
TUEP & AUROC$^{p}$ & yes, all & 0.679 {\scriptsize[0.607, 0.751]} & 0.658 {\scriptsize[0.583, 0.739]} & \textbf{0.694} {\scriptsize[0.624, 0.766]} \\
TUEV & BAcc & yes, all$^{b}$ & \textbf{0.542} {\scriptsize[0.529, 0.554]} & 0.450$\downarrow$ {\scriptsize[0.437, 0.461]} & \textbf{0.542} {\scriptsize[0.529, 0.554]} \\
EEGMMIDB & BAcc & no & \textbf{0.442} {\scriptsize[0.419, 0.463]} & 0.326$\downarrow$ {\scriptsize[0.303, 0.349]} & 0.361$\downarrow$ {\scriptsize[0.337, 0.383]} \\
Sleep-EDF & BAcc & other subjects & \textbf{0.757} {\scriptsize[0.748, 0.766]} & 0.721$\downarrow$ {\scriptsize[0.711, 0.732]} & 0.724$\downarrow$ {\scriptsize[0.714, 0.733]} \\
SEED-V & BAcc & yes, all & \textbf{0.315} {\scriptsize[0.310, 0.319]} & 0.296$\downarrow$ {\scriptsize[0.291, 0.300]} & 0.301$\downarrow$ {\scriptsize[0.297, 0.306]} \\
\bottomrule
\end{tabular}
\end{table}

\subsection{Model interpretation}
\label{sec:attribution}

\begin{figure}[!htbp]
  \centering
  \includegraphics[width=\textwidth]{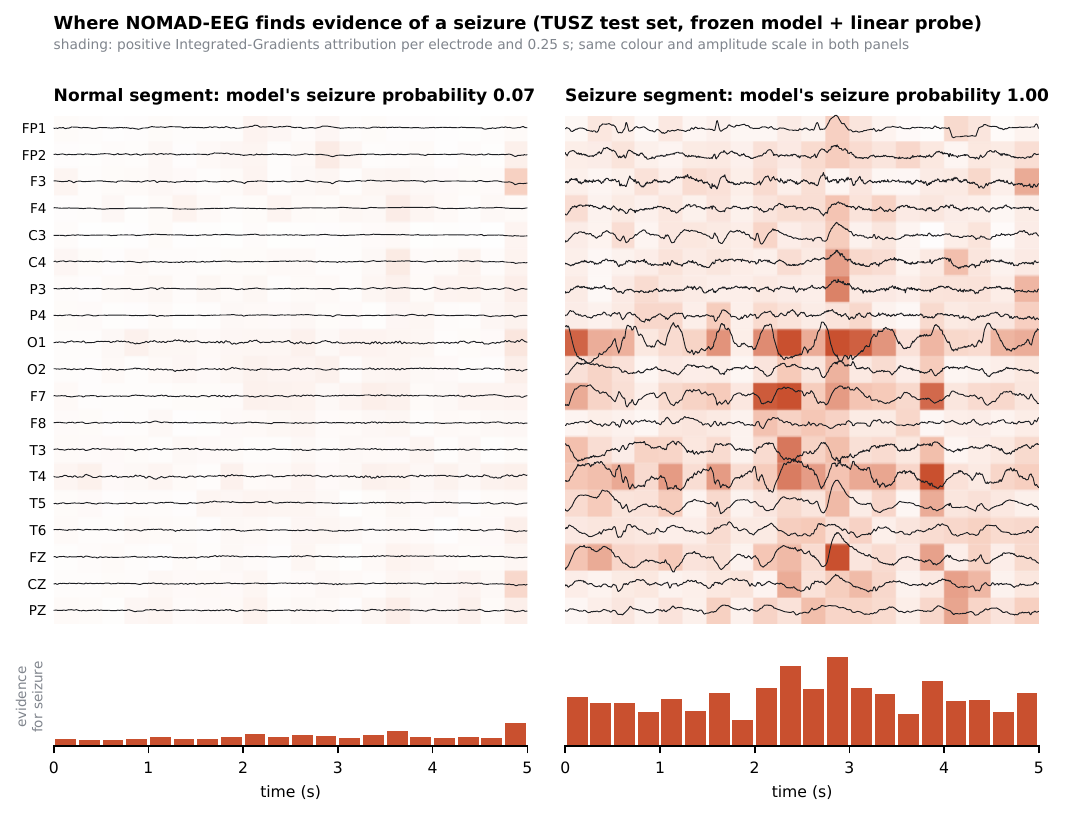}
  \caption{\textbf{Where \model{} finds evidence of a seizure.} Two 5\,s segments from the TUSZ test
  set, scored by the frozen \model{} with the TUSZ linear probe of \cref{tab:tusz}. The segments were
  chosen by a rule fixed in advance: a random (seeded) test window labelled seizure that the model
  classifies with probability above 0.9, and a random normal window classified with probability
  below 0.1, i.e.\ typical correct decisions rather than hand-picked ones. Shading behind each trace:
  positive Integrated-Gradients attribution (evidence for seizure) per electrode and 0.25\,s; bottom:
  its sum over electrodes. Both panels share the colour and amplitude scales. The figure illustrates
  single decisions and is not a quantitative test of seizure localisation.}
  \label{fig:attribution}
\end{figure}

In the seizure segment the evidence for seizure concentrates on the sharp, high-amplitude
discharges, and its time course peaks where they occur; in the normal segment there is almost no
evidence anywhere (\cref{fig:attribution}).

\section{Discussion}
\label{sec:discussion}
\subsection{Principal findings}
\label{sec:disc_main}

\model{} was designed to read any montage and any recording length with one set of weights, and the
experiments test that design rather than a single benchmark. The same 12.0\,M-parameter model, without
any architectural change, read one electrode (Bonn), two bipolar derivations (Sleep-EDF), eight
(BCIAUT-P300), 19 (Mumtaz), 21 (TUH corpora) and 64 electrodes (EEGMMIDB), and segments from 1.4\,s to
30\,s. Across eight benchmarks it ranks never below fourth among eight foundation models, which places a
model pre-trained on 147\,h of EEG among models trained on far more data: REVE-base has 69\,M
parameters and was pre-trained on about 60{,}000\,h. It is the best model where the montage departs
most from its pre-training: on EEGMMIDB, where most electrode positions are new to it (and which LaBraM
saw in pre-training, \cref{sec:baselines}), and on BCIAUT-P300, a new population, task and
eight-electrode montage, where it detects the attended object in 75.5\% of blocks against 61.2\% for
the next-best model. We read this as evidence that identifying electrodes by position in every block,
rather than by name or slot, transfers to unseen layouts. Reading the single Bonn electrode on its
own, which still gives an AUROC of 0.995, is an input that no fixed-montage model can be given. On
Siena, a seizure corpus external to \model{} (but part of REVE's, LaBraM's and EEGMamba's pre-training
data), \model{} is the most precise detector: with 99.3\% of test windows free of seizures, its highest
accuracy, weighted F1 and macro F1, the last with a confidence interval that excludes every other
model, mean the fewest false alarms at the decision threshold.

\subsection{Where \model{} trails, and why}
\label{sec:disc_trails}

REVE and EEGDM lead on TUAB, where BIOT is also ahead of \model{} (AUROC 0.847 against 0.838); on
TUSZ's AUROC, PR-AUC and balanced accuracy, where CBraMod also has a higher PR-AUC (0.455 against
0.446); on Sleep-EDF; and on Mumtaz, where LaBraM also has a higher balanced accuracy (0.838 against
0.832). On TUEV only EEGDM's balanced accuracy exceeds fine-tuned \model{}, which is best on $\kappa$
and weighted F1. EEGDM's TUEV results should be read in light of its pre-training, which, like ours,
includes TUEV recordings. On TUEP EEGMamba has a higher patient-level AUROC (0.694 against 0.679),
although the paired difference is not significant. On Siena EEGDM and REVE rank seizure windows better
than \model{} (AUROC 0.915 and 0.906 against 0.882; REVE also has the highest PR-AUC), so \model{}'s
Siena advantage lies in fewer false alarms rather than in ranking. Two factors are plausible. First,
scale: REVE-base has about six times as many parameters as \model{} and was pre-trained on roughly 400
times as much EEG. Second, domain (\cref{sec:baselines}): REVE, CBraMod and EEGMamba were pre-trained on
the TUH corpus from which TUAB, TUSZ, TUEV and TUEP are drawn (REVE without the TUAB and TUEV
recordings), BIOT's checkpoint used the TUAB and TUEV training splits, EEGDM was pre-trained on TUEV,
LaBraM on TUSZ and TUEP, and REVE saw Siena, whereas \model{} saw only TUEV. TUEP illustrates the second
factor: the only compared model with a higher AUROC there was pre-trained on the TUH corpus that
contains TUEP, and among the models pre-trained neither on TUEP nor on the full TUH corpus, \model{} is
best on every metric. Scale alone did not close this gap in our hands: \model{}-Large, with 5.6 times
the parameters and 47 times the pre-training data, including the TUH corpora, was not significantly
better than Base on any benchmark; its point estimate was higher only on Bonn and TUEP, with
overlapping intervals, equal on TUEV and lower on the other seven (\cref{sec:disc_scale}).

\subsection{Transfer to unseen populations}
\label{sec:disc_transfer}

On BCIAUT-P300, BIOT and DiffEEG are close to chance. We attribute this mainly to how little of the
montage they can read: two bipolar pairs for BIOT, and six of eight electrodes for DiffEEG. Bipolar
derivations also cancel part of a P300 that is broad and similar across these closely spaced
electrodes. On Parkinson's disease \model{} is not the best model, and EEGDM's lead is significant.
\model{}'s selected probe strength varied by five orders of magnitude across folds, a sign that its
probe is unstable with about 24 validation participants per fold. One possible reason for the gap,
which we have not tested, is that the resting-state changes reported in PD are spectral properties of
the whole recording rather than transient events like those in \model{}'s pre-training data.

\subsection{Pre-training, objective and fine-tuning}
\label{sec:disc_training}

The random-weights control shows that pre-training adds a large, significant gain on TUEV, TUAB and
Siena, but also that the architecture with random weights is already competitive on small or
saturated tasks: on Mumtaz, fine-tuning from random weights matches fine-tuning from the pre-trained
model, which suggests that this two-class task is solvable by the architecture alone, and Bonn is
saturated for both. Results on such tasks should therefore not be credited to pre-training without
this control.

In a controlled comparison, denoising diffusion gave better frozen representations than flow matching
on four datasets and worse on one, so we use diffusion for \model{}. This does not contradict reports
that flow matching pre-trains EEG models efficiently~\citep{hwang2026bfm}, which used a different
architecture, more data and fine-tuning; it shows that the choice of objective matters and should be
tested rather than assumed for a given backbone.

End-to-end fine-tuning helped on TUEV, EEGMMIDB and SEED-V but not on Mumtaz, Sleep-EDF, TUAB or
TUSZ, where the frozen probe was as good or better. In both failure cases the choice of the
fine-tuning epoch is the weak point. On Mumtaz, validation accuracy approached 1.0 while test
accuracy stayed near 0.82, a sign that a validation set of nine subjects cannot pick a checkpoint
that generalises to new people. On TUAB and TUSZ the validation windows come from the training
patients themselves (\cref{sec:finetune}), and validation balanced accuracy reached 0.98 and 0.96
against test values of 0.77, so the selection rewarded fitting those patients rather than
generalising to new ones. On SEED-V, by contrast, the validation clips are distinct from the training
clips. Selecting checkpoints on enough held-out subjects, or by cross-validation over subjects, is
therefore a precondition for a fair fine-tuning comparison.

\subsection{Electrode order and missing electrodes}
\label{sec:missing}

The order test confirms in practice what \cref{sec:properties} proves, and the position-shuffling
control shows that the invariance does not come from ignoring spatial information. Removing electrodes
leaves the probe's ranking of test windows largely intact but shifts the scores relative to the
threshold learned on the full montage, because the probe reads the average over electrodes; this is
why balanced accuracy falls faster than AUROC. A threshold recalibrated on the reduced montage, or a
probe trained with electrodes dropped at random, would address this. Base was pre-trained on a single
21-electrode montage and never saw a reduced one, which is one reason \model{}-Large was pre-trained on
random electrode subsets of 2 to 32 electrodes.

\subsection{Scaling model size and pre-training data}
\label{sec:disc_scale}

Four findings follow from \cref{tab:scaling}. First, scaling model and data together did not improve
the frozen features: with 5.6 times the parameters and 47 times the pre-training data,
\model{}-Large was not significantly better than Base on any benchmark after 1.9 epochs and was
clearly worse on three. Its point estimate was higher than Base's on only two datasets, Bonn (AUROC
1.000 against 0.999, where both are at the ceiling) and TUEP (0.694 against 0.679, paired difference
not significant), and after 1.0 epoch on TUSZ (0.878 against 0.866, overlapping intervals). Second, the largest losses are on the tasks least like clinical EEG: motor imagery
(EEGMMIDB, $-0.081$ balanced accuracy), sleep staging (Sleep-EDF, $-0.033$), emotion (SEED-V,
$-0.014$) and depression (Mumtaz, $-0.029$ AUROC), although Large was pre-trained on many montages, on
random electrode subsets and, for SEED-V and Sleep-EDF, on the same corpora. Third, seeing the
evaluation corpus in pre-training did not help: SEED-V and TUEP were in Large's pre-training data and
not in Base's, yet Large is below Base on SEED-V and level on TUEP. Fourth, longer training did not
improve every task: from 1.0 to 1.9 epochs TUEV and EEGMMIDB improved while TUSZ dropped from
above Base's point estimate to below it, so a single intermediate checkpoint can over- or understate a
model.

These results agree with recent EEG studies in which larger models or more pre-training data did not
reliably improve downstream performance. Across more than 20{,}000 evaluations of 30 EEG foundation
models, parameter count showed no consistent positive relationship with downstream
performance~\citep{chen2026eegarena}. EEG-FM-Bench found that model or data scale alone does not
explain transfer, and pointed to the pre-training objective and EEG-specific design
instead~\citep{xiong2026eegfmbench}. Fine-tuned large brainwave models improved on much smaller
conventional networks by only about one percentage point while using far more
parameters~\citep{lee2025lbm}. In a channel-adaptation benchmark the 5\,M-parameter CBraMod
outperformed models up to 31 times larger on four of five datasets~\citep{channeladapt2026}. HEAR's
base model gained little over its tiny variant, which its authors attributed to saturation of the
available pre-training data~\citep{chen2025hear}, and CodeBrain reported consistent but diminishing
returns with model size~\citep{ma2026codebrain}. The evidence on data is more mixed: under a fixed
architecture, more pre-training data gave sustained gains in \citet{chen2026eegarena}.

That study held the architecture fixed while varying data, whereas we changed model size and data
together on a fixed training budget. Large saw 1.63\,M samples against Base's 1.28\,M, at most about two
epochs over its pool against about 74 for Base over TUEV (\cref{tab:scaling_setup}), so it may be
under-trained relative to its data, and its features are twice as wide for the same linear probe. We
have not tested which of these explains the result. Our result is therefore evidence that scale does
not help this architecture without a matching increase in compute, not that scale cannot help; a study
that scales compute with model and data, and that varies model size and data size separately, is
needed to settle it.

\subsection{What the model responds to}
\label{sec:disc_attribution}

In the attribution example, the evidence for seizure falls on the sharp, high-amplitude discharges
that a clinician would point to, and almost none falls on the normal segment. The two segments were
chosen by a rule fixed in advance, but they illustrate single decisions: they show that the probe's
decision can be traced to plausible parts of the signal, not that \model{} localises seizures, which
would require a quantitative comparison with annotated onsets and channels.

\subsection{Limitations}
\label{sec:limitations}

\begin{itemize}
  \item \textbf{Single pre-training corpus.} \model{} was pre-trained on 147\,h of TUEV only, with one
  montage. Positions outside that montage rely on the smoothness of the positional features. The
  larger multi-corpus model was not significantly better on any benchmark (\cref{sec:disc_scale}),
  but it changed model size
  and data together, on one GPU slice and without a matching increase in training compute.
  \item \textbf{Compatibility is not robustness.} Accepting any montage does not guarantee equal
  accuracy on every montage~\citep{steglich2026montage}. Channel order does not affect \model{},
  and with electrodes removed it still ranks examples well, but the decision threshold of a probe
  fitted on the full montage no longer fits (\cref{sec:missing}).
  \item \textbf{Inputs.} The model assumes 200\,Hz input and a known 3D position for every channel;
  channels without a standard position are dropped, and a bipolar derivation is approximated by a
  sensor at its midpoint.
  \item \textbf{Context length.} The architecture accepts inputs longer than its 30\,s pre-training
  window, but we have not yet evaluated inputs longer than 30\,s.
  \item \textbf{Evaluation.} Baselines were evaluated only with frozen linear probes; each with its
  own pooling, so their probe input sizes differ (200 to 10{,}240 dimensions). Fine-tuned comparisons
  with the baselines remain to be run.
\end{itemize}

\subsection{Conclusion}
\label{sec:conclusion}

\model{} shows that a diffusion-pretrained EEG model can identify electrodes only by their 3D
position and model time without patches, and still be competitive with foundation models that are
larger and trained on far more data. With one set of weights pre-trained on 147\,h of one corpus, it
read montages from one to 64 electrodes and was never ranked below fourth on eight benchmarks. It was
the best model on the layouts furthest from its pre-training (EEGMMIDB and BCIAUT-P300) and on three
epilepsy tasks: seizure detection on Siena (macro F1) and Bonn, and epilepsy diagnosis on TUEP among
the models not pre-trained on it; after fine-tuning it also had the highest $\kappa$ and weighted F1 on
TUEV event classification. Scaling the model and its data together, without scaling compute, did
not significantly improve it on any benchmark. Pre-training on more
diverse montages with a compute budget that grows with the data, and recalibrating task heads for
reduced montages, are the most direct next steps.

\section*{Declarations}
\noindent\textbf{Funding.} {This work was supported by the LAU President's Intramural Research Fund- I0084}

\noindent\textbf{Competing interests.} The authors declare no conflict of interest.

\noindent\textbf{Ethics approval.} This study used only publicly available, de-identified EEG datasets
(TUH EEG corpora, Siena Scalp EEG, Bonn, Mumtaz, EEGMMIDB, Sleep-EDF, SEED-V, BCIAUT-P300 and OpenNeuro
ds004584), each collected under the ethics approval stated by its original authors.

\noindent\textbf{Consent to participate.} Not applicable; informed consent was obtained by the
original data collectors.

\noindent\textbf{Data availability.} All datasets are publicly available from their original sources

\noindent\textbf{Code availability.} Repository link for code and model weights will be shared once the manuscript is published.

\bibliographystyle{unsrtnat}
\bibliography{refs}

@article{gu2022s4,
  title={{Efficiently Modeling Long Sequences with Structured State Spaces}},
  author={Albert Gu and Karan Goel and Christopher Ré},
  journal={arXiv preprint arXiv:2111.00396},
  year={2021}
}

@article{gu2022s4d,
  title={{On the Parameterization and Initialization of Diagonal State Space Models}},
  author={Albert Gu and Ankit Gupta and Karan Goel and Christopher Ré},
  journal={arXiv preprint arXiv:2206.11893},
  year={2022}
}

@article{ho2020ddpm,
  title={{Denoising Diffusion Probabilistic Models}},
  author={Jonathan Ho and Ajay Jain and Pieter Abbeel},
  journal={arXiv preprint arXiv:2006.11239},
  year={2020}
}

@article{nichol2021iddpm,
  title={{Improved Denoising Diffusion Probabilistic Models}},
  author={Alex Nichol and Prafulla Dhariwal},
  journal={arXiv preprint arXiv:2102.09672},
  year={2021}
}

@article{lipman2023flow,
  title={{Flow Matching for Generative Modeling}},
  author={Yaron Lipman and Ricky T. Q. Chen and Heli Ben-Hamu and Maximilian Nickel and Matt Le},
  journal={arXiv preprint arXiv:2210.02747},
  year={2022}
}

@article{liu2023rectified,
  title={{Flow Straight and Fast: Learning to Generate and Transfer Data with Rectified Flow}},
  author={Xingchao Liu and Chengyue Gong and Qiang Liu},
  journal={arXiv preprint arXiv:2209.03003},
  year={2022}
}

@article{oord2016wavenet,
  title={{WaveNet: A Generative Model for Raw Audio}},
  author={Aaron van den Oord and Sander Dieleman and Heiga Zen and Karen Simonyan and Oriol Vinyals and Alex Graves and Nal Kalchbrenner and Andrew Senior and Koray Kavukcuoglu},
  journal={arXiv preprint arXiv:1609.03499},
  year={2016}
}

@article{perez2018film,
  title={{FiLM: Visual Reasoning with a General Conditioning Layer}},
  author={Ethan Perez and Florian Strub and Harm de Vries and Vincent Dumoulin and Aaron Courville},
  journal={arXiv preprint arXiv:1709.07871},
  year={2017}
}

@article{tancik2020fourier,
  title={{Fourier Features Let Networks Learn High Frequency Functions in Low Dimensional Domains}},
  author={Matthew Tancik and Pratul P. Srinivasan and Ben Mildenhall and Sara Fridovich-Keil and Nithin Raghavan and Utkarsh Singhal and Ravi Ramamoorthi and Jonathan T. Barron and Ren Ng},
  journal={arXiv preprint arXiv:2006.10739},
  year={2020}
}

@article{mildenhall2020nerf,
  title={{NeRF: Representing Scenes as Neural Radiance Fields for View Synthesis}},
  author={Ben Mildenhall and Pratul P. Srinivasan and Matthew Tancik and Jonathan T. Barron and Ravi Ramamoorthi and Ren Ng},
  journal={arXiv preprint arXiv:2003.08934},
  year={2020}
}

@article{gu2023mamba,
  title={{Mamba: Linear-Time Sequence Modeling with Selective State Spaces}},
  author={Albert Gu and Tri Dao},
  journal={arXiv preprint arXiv:2312.00752},
  year={2023}
}

@article{vaswani2017attention,
  title={{Attention Is All You Need}},
  author={Ashish Vaswani and Noam Shazeer and Niki Parmar and Jakob Uszkoreit and Llion Jones and Aidan N. Gomez and Lukasz Kaiser and Illia Polosukhin},
  journal={arXiv preprint arXiv:1706.03762},
  year={2017}
}

@article{jiang2024labram,
  title={{Large Brain Model for Learning Generic Representations with Tremendous EEG Data in BCI}},
  author={Wei-Bang Jiang and Li-Ming Zhao and Bao-Liang Lu},
  journal={arXiv preprint arXiv:2405.18765},
  year={2024}
}

@article{yang2023biot,
  title={{BIOT: Cross-data Biosignal Learning in the Wild}},
  author={Chaoqi Yang and M. Brandon Westover and Jimeng Sun},
  journal={arXiv preprint arXiv:2305.10351},
  year={2023}
}

@article{wang2025cbramod,
  title={{CBraMod: A Criss-Cross Brain Foundation Model for EEG Decoding}},
  author={Jiquan Wang and Sha Zhao and Zhiling Luo and Yangxuan Zhou and Haiteng Jiang and Shijian Li and Tao Li and Gang Pan},
  journal={arXiv preprint arXiv:2412.07236},
  year={2024}
}

@article{puah2025eegdm,
  title={{EEGDM: EEG Representation Learning via Generative Diffusion Model}},
  author={Jia Hong Puah and Sim Kuan Goh and Ziwei Zhang and Zixuan Ye and Chow Khuen Chan and Kheng Seang Lim and Si Lei Fong and Kok Sin Woon and Cuntai Guan},
  journal={arXiv preprint arXiv:2508.14086},
  year={2025}
}

@article{diffeeg2026,
  title={{DiffEEG: A Self-Supervised Denoising Diffusion Model for Learning EEG Generic Representations}},
  author={Abdulkader Helwan and Lina Abou-Abbas and Hussein El Amouri and Belkacem Chikhaoui and Khadidja Henni},
  journal={arXiv preprint arXiv:2607.11578},
  year={2026}
}

@article{hwang2026bfm,
  title={{B[FM]$^2$: Brain Foundation Model via Flow Matching with SplitUNet}},
  author={Jaedong Hwang and Kathleen Zhang and Wei Dai and Konstantinos Kontras and Maarten Vanmarcke and Maarten De Vos and Ila Fiete and Paul Pu Liang},
  journal={arXiv preprint arXiv:2606.20812},
  year={2026}
}

@article{ouahidi2025reve,
  title={{REVE: A Foundation Model for EEG -- Adapting to Any Setup with Large-Scale Pretraining on 25,000 Subjects}},
  author={Yassine El Ouahidi and Jonathan Lys and Philipp Thölke and Nicolas Farrugia and Bastien Pasdeloup and Vincent Gripon and Karim Jerbi and Giulia Lioi},
  journal={arXiv preprint arXiv:2510.21585},
  year={2025}
}

@article{doner2025luna,
  title={{LUNA: Efficient and Topology-Agnostic Foundation Model for EEG Signal Analysis}},
  author={Berkay Döner and Thorir Mar Ingolfsson and Luca Benini and Yawei Li},
  journal={arXiv preprint arXiv:2510.22257},
  year={2025}
}

@article{xiao2025brainomni,
  title={{BrainOmni: A Brain Foundation Model for Unified EEG and MEG Signals}},
  author={Qinfan Xiao and Ziyun Cui and Chi Zhang and Siqi Chen and Wen Wu and Andrew Thwaites and Alexandra Woolgar and Bowen Zhou and Chao Zhang},
  journal={arXiv preprint arXiv:2505.18185},
  year={2025}
}

@article{han2025diver0,
  title={{DIVER-0 : A Fully Channel Equivariant EEG Foundation Model}},
  author={Danny Dongyeop Han and Ahhyun Lucy Lee and Taeyang Lee and Yonghyeon Gwon and Sebin Lee and Seongjin Lee and David Keetae Park and Shinjae Yoo and Jiook Cha and Chun Kee Chung},
  journal={arXiv preprint arXiv:2507.14141},
  year={2025}
}

@article{steglich2026montage,
  title={{Representational and Functional Robustness to Electrode Montages in EEG Foundation Models}},
  author={Jakob Steglich and Justus Meyer zu Bexten and Shakiba Moradi and Laure Ciernik and Simon M. Hofmann and Mina Jamshidi Idaji},
  journal={arXiv preprint arXiv:2609.36288},
  year={2026}
}

@article{tgsd2026,
  title={{TGSD: Topology-Guided State-Space Diffusion Framework for EEG Spatial Super-Resolution}},
  author={Zijian Kang and Weiming Zeng and Yueyang Li and Shengyu Gong and Hongjie Yan and Wai Ting Siok and Nizhuan Wang},
  journal={arXiv preprint arXiv:2606.03998},
  year={2026}
}

@article{kostas2021bendr,
  title={{BENDR: using transformers and a contrastive self-supervised learning task to learn from massive amounts of EEG data}},
  author={Demetres Kostas and Stephane Aroca-Ouellette and Frank Rudzicz},
  journal={arXiv preprint arXiv:2101.12037},
  year={2021}
}

@article{kuruppu2025review,
  title={{EEG Foundation Models: A Critical Review of Current Progress and Future Directions}},
  author={Gayal Kuruppu and Neeraj Wagh and Vaclav Kremen and Sandipan Pati and Gregory Worrell and Yogatheesan Varatharajah},
  journal={arXiv preprint arXiv:2507.11783},
  year={2025}
}

@article{channeladapt2026,
  title={{Channel Adaptation for EEG Foundation Models: A Systematic Benchmark Across Architectures, Tasks, and Training Regimes}},
  author={Kuntal Kokate and Bruno Aristimunha and Dung Truong and Arnaud Delorme},
  journal={arXiv preprint arXiv:2604.23091},
  year={2026}
}

@article{loshchilov2019adamw,
  title={{Decoupled Weight Decay Regularization}},
  author={Ilya Loshchilov and Frank Hutter},
  journal={arXiv preprint arXiv:1711.05101},
  year={2017}
}

@article{chen2016checkpoint,
  title={{Training Deep Nets with Sublinear Memory Cost}},
  author={Tianqi Chen and Bing Xu and Chiyuan Zhang and Carlos Guestrin},
  journal={arXiv preprint arXiv:1604.06174},
  year={2016}
}

@article{zhou2025survey,
  title={{Brain Foundation Models: A Survey on Advancements in Neural Signal Processing and Brain Discovery}},
  author={Xinliang Zhou and Chenyu Liu and Zhisheng Chen and Kun Wang and Yi Ding and Ziyu Jia and Qingsong Wen},
  journal={arXiv preprint arXiv:2503.00580},
  year={2025}
}

@article{obeid2016tuh,
  title={{The Temple University Hospital EEG Data Corpus}},
  author={Iyad Obeid and Joseph Picone},
  journal={Frontiers in Neuroscience},
  year={2016},
  doi={10.3389/fnins.2016.00196}
}

@article{harati2015tuev,
  title={{Improved EEG event classification using differential energy}},
  author={A. Harati and M. Golmohammadi and S. Lopez and I. Obeid and J. Picone},
  journal={2015 IEEE Signal Processing in Medicine and Biology Symposium (SPMB)},
  year={2015},
  doi={10.1109/spmb.2015.7405421}
}

@article{gramfort2013mne,
  title={{MEG and EEG data analysis with MNE-Python}},
  author={Alexandre Gramfort},
  journal={Frontiers in Neuroscience},
  year={2013},
  doi={10.3389/fnins.2013.00267}
}

@article{wang2024eegpt,
  title={{EEGPT: Pretrained Transformer for Universal and Reliable Representation of EEG Signals}},
  author={Guagnyu Wang and Wenchao Liu and Yuhong He and Cong Xu and Lin Ma and Haifeng Li},
  journal={Advances in Neural Information Processing Systems 37},
  year={2024},
  doi={10.52202/079017-1239}
}

@article{yi2023mmm,
  title={{Learning Topology-Agnostic EEG Representations with Geometry-Aware Modeling}},
  author={Ke Yi and Yansen Wang and Kan Ren and Dongsheng Li},
  journal={Advances in Neural Information Processing Systems 36},
  year={2023},
  doi={10.52202/075280-2344}
}

@article{wang2025eegmamba,
  title={{EEGMamba: An EEG foundation model with Mamba}},
  author={Jiquan Wang and Sha Zhao and Zhiling Luo and Yangxuan Zhou and Shijian Li and Gang Pan},
  journal={Neural Networks},
  year={2025},
  doi={10.1016/j.neunet.2025.107816}
}

@article{kane2017glossary,
  title={{A revised glossary of terms most commonly used by clinical electroencephalographers and updated proposal for the report format of the EEG findings. Revision 2017}},
  author={Nick Kane and Jayant Acharya and Sandor Beniczky and Luis Caboclo and Simon Finnigan and Peter W. Kaplan and Hiroshi Shibasaki and Ronit Pressler and Michel J.A.M. van Putten},
  journal={Clinical Neurophysiology Practice},
  year={2017},
  doi={10.1016/j.cnp.2017.07.002}
}

@article{simoes2020bciaut,
  title={{BCIAUT-P300: A Multi-Session and Multi-Subject Benchmark Dataset on Autism for P300-Based Brain-Computer-Interfaces}},
  author={Marco Simões and Davide Borra and Eduardo Santamaría-Vázquez and {GBT-UPM} and Mayra Bittencourt-Villalpando and Dominik Krzemiński and Aleksandar Miladinović and {Neural Engineering Group} and Thomas Schmid and Haifeng Zhao and Carlos Amaral and Bruno Direito and Jorge Henriques and Paulo Carvalho and Miguel Castelo-Branco},
  journal={Frontiers in Neuroscience},
  year={2020},
  doi={10.3389/fnins.2020.568104}
}

@article{chen2026eegarena,
  title={{Benchmarking EEG Foundation Models at Scale: Lessons from 20,000 Evaluations}},
  author={Zhige Chen and Shu Peng and Chengxuan Qin and Rui Liu and Rui Yang and Kay Chen Tan and Jibin Wu},
  journal={arXiv preprint arXiv:2609.32743},
  year={2026}
}

@inproceedings{xiong2026eegfmbench,
  title={{EEG-FM-Bench: A Comprehensive Benchmark for the Systematic Evaluation and Diagnostic Analyses of EEG Foundation Models}},
  author={Wei Xiong and Jiangtong Li and Jie Li and Kun Zhu and Changjun Jiang},
  booktitle={International Conference on Machine Learning (ICML)},
  year={2026},
  note={arXiv:2508.17742}
}

@inproceedings{lee2025lbm,
  title={{Are Large Brainwave Foundation Models Capable Yet? Insights from Fine-tuning}},
  author={Na Lee and Konstantinos Barmpas and Yannis Panagakis and Dimitrios Adamos and Nikolaos Laskaris and Stefanos Zafeiriou},
  booktitle={International Conference on Machine Learning (ICML)},
  year={2025},
  note={arXiv:2507.01196}
}

@article{chen2025hear,
  title={{HEAR: An EEG Foundation Model with Heterogeneous Electrode Adaptive Representation}},
  author={Zhige Chen and Chengxuan Qin and Wenlong You and Rui Liu and Congying Chu and Rui Yang and Kay Chen Tan and Jibin Wu},
  journal={arXiv preprint arXiv:2510.12515},
  year={2025}
}

@inproceedings{ma2026codebrain,
  title={{CodeBrain: Bridging Decoupled Tokenizer and Multi-Scale Architecture for EEG Foundation Model}},
  author={Jingying Ma and Feng Wu and Qika Lin and Yucheng Xing and Chenyu Liu and Ziyu Jia and Mengling Feng},
  booktitle={International Conference on Learning Representations (ICLR)},
  year={2026},
  note={arXiv:2506.09110}
}

@inproceedings{sundararajan2017ig,
  title={{Axiomatic Attribution for Deep Networks}},
  author={Mukund Sundararajan and Ankur Taly and Qiqi Yan},
  booktitle={International Conference on Machine Learning (ICML)},
  pages={3319--3328},
  year={2017}
}
\end{document}